\documentclass{article}

\usepackage{PRIMEarxiv}

\usepackage[utf8]{inputenc} 
\usepackage[T1]{fontenc}    
\usepackage{hyperref}       
\usepackage{url}            
\usepackage{booktabs}       
\usepackage{amsfonts}       
\usepackage{nicefrac}       
\usepackage{microtype}      
\usepackage{lipsum}
\usepackage{fancyhdr}       
\usepackage{graphicx}       
\graphicspath{{media/}}     

\usepackage{multirow}
\usepackage[table]{xcolor}
\usepackage{float}
\usepackage{tcolorbox}
\tcbuselibrary{breakable}
\usepackage{listings}
\usepackage{longtable}
\usepackage{array}

\usepackage{pifont}
\usepackage{amsmath}

\newcommand{\cmark}{\ding{51}}
\newcommand{\xmark}{\ding{55}}

\newtcolorbox{promptbox}{
    colback=gray!5,
    colframe=gray!60,
    boxrule=0.6pt,
    arc=2pt,
    left=6pt,
    right=6pt,
    top=6pt,
    bottom=6pt,
    breakable
}

\definecolor{topgreen}{RGB}{220,242,220}
\definecolor{topblue}{RGB}{217,233,248}
\definecolor{topyellow}{RGB}{247,241,204}

\newcommand{\rankone}[1]{%
  \begingroup
  \setlength{\fboxsep}{1pt}%
  \colorbox{topgreen}{#1}%
  \endgroup
}
\newcommand{\ranktwo}[1]{%
  \begingroup
  \setlength{\fboxsep}{1pt}%
  \colorbox{topblue}{#1}%
  \endgroup
}
\newcommand{\rankthree}[1]{%
  \begingroup
  \setlength{\fboxsep}{1pt}%
  \colorbox{topyellow}{#1}%
  \endgroup
}

\title{UltraG-Bench: A Multi-task Benchmark for assessing Large Vision-Language Models on Pixel-level Evidence Grounding in Ultrasound}

\author{
  Quanhao Zhu, Bo Xu, Rui Lin, Chenyuan Wang, Yu Shao, Boling Zhu, Jiuyan Sun, Liang Zhao, Hongfei Lin \\
  Dalian University of Technology \\
  \texttt{zhuqh19@gmail.com} \\
  \texttt{boxu@dlut.edu.cn} \\
  \texttt{lr983088162@mail.dlut.edu.cn} \\
  \texttt{chenyuanwang@mail.dlut.edu.cn} \\
  \texttt{python@mail.dlut.edu.cn} \\
  \texttt{SinsapZ@outlook.com} \\
  \texttt{2093145195@mail.dlut.edu.cn} \\
  \texttt{liangzhao@dlut.edu.cn} \\
  \texttt{hflin@dlut.edu.cn}
  \AND
  Feng Xia \\
  RMIT University \\
  \texttt{f.xia@ieee.org}
}

\begin{document}
\maketitle

\begin{abstract}
Ultrasound is one of the most widely used medical imaging modalities, and recent large vision-language models(VLMs) have shown increasing capabilities in ultrasound image understanding. However, these models fail to provide pixel-level visual evidence aligned with their semantic predictions, and their fine-grained grounding capability in ultrasound remains largely unclear. We introduce UltraG-Bench, a large-scale multi-task benchmark for evaluating pixel-level evidence grounding in ultrasound. UltraG-Bench is built by annotating 40 public ultrasound segmentation datasets spanning 13 anatomical categories, and comprises three progressive tasks: instruction-guided segmentation, evidence-grounded VQA, and evidence-grounded report generation, with 331125, 666779, and 138832 annotations, respectively. Comprehensive evaluation of 14 state-of-the-art models reveals a substantial gap between semantic understanding and fine-grained pixel-level localization. We further propose UltraG-Agent, which combines the semantic reasoning capabilities of a VLM with the ultrasound-specific segmentation capability of UltraSAM3. Experiments show that UltraG-Agent substantially improves both semantic prediction and pixel-level visual grounding. Our dataset and code are available at \url{https://github.com/zhuqh19/UltraG-Bench}.
\end{abstract}


\section{Introduction}
Ultrasound has become one of the most widely used medical imaging modalities in clinical practice owing to its lack of ionizing radiation, real-time imaging capability, and relatively low cost. Compared with CT and MRI, ultrasound devices are more portable and less expensive, making them highly accessible across diverse clinical settings~\cite{fleming2021lancet}. Accurate recognition, localization, and interpretation of anatomical categories and lesions in ultrasound images are fundamental to clinical diagnosis and decision-making~\cite{guo2026visually}. 

Nowadays, supervised segmentation methods and general foundation models have continuously improved pixel-level localization of organs and lesions~\cite{ronneberger2015u,ma2024segment}. On the other hand, vision-language models (VLMs) and multimodal large language models (MLLMs) are increasingly being applied to ultrasound image understanding, including tasks such as visual question answering (VQA) and report generation~\cite{guo2026visually}. However, a substantial gap remains between pixel-level visual localization and high-level semantic understanding in current models. Conventional segmentation models can provide precise spatial boundaries but generally lack open-ended language understanding and semantic generation capabilities, whereas VLMs and MLLMs can answer questions or generate image descriptions but often fail to provide pixel-level visual evidence that is aligned with their semantic predictions~\cite{myhre2026artificial,jin2026grounded,xu2025echogpt,xumedvcot}. This gap is particularly critical in medical applications, where clinical reasoning depends not only on “what the model predicts”, but also on “which regions in the image support that prediction.” However, this gap has not yet been systematically evaluated by existing benchmarks, leaving the extent of the disconnect between semantic understanding and pixel-level visual grounding in current models poorly understood.

\begin{figure}[!ht]
    \centering
    \includegraphics[width=\textwidth]{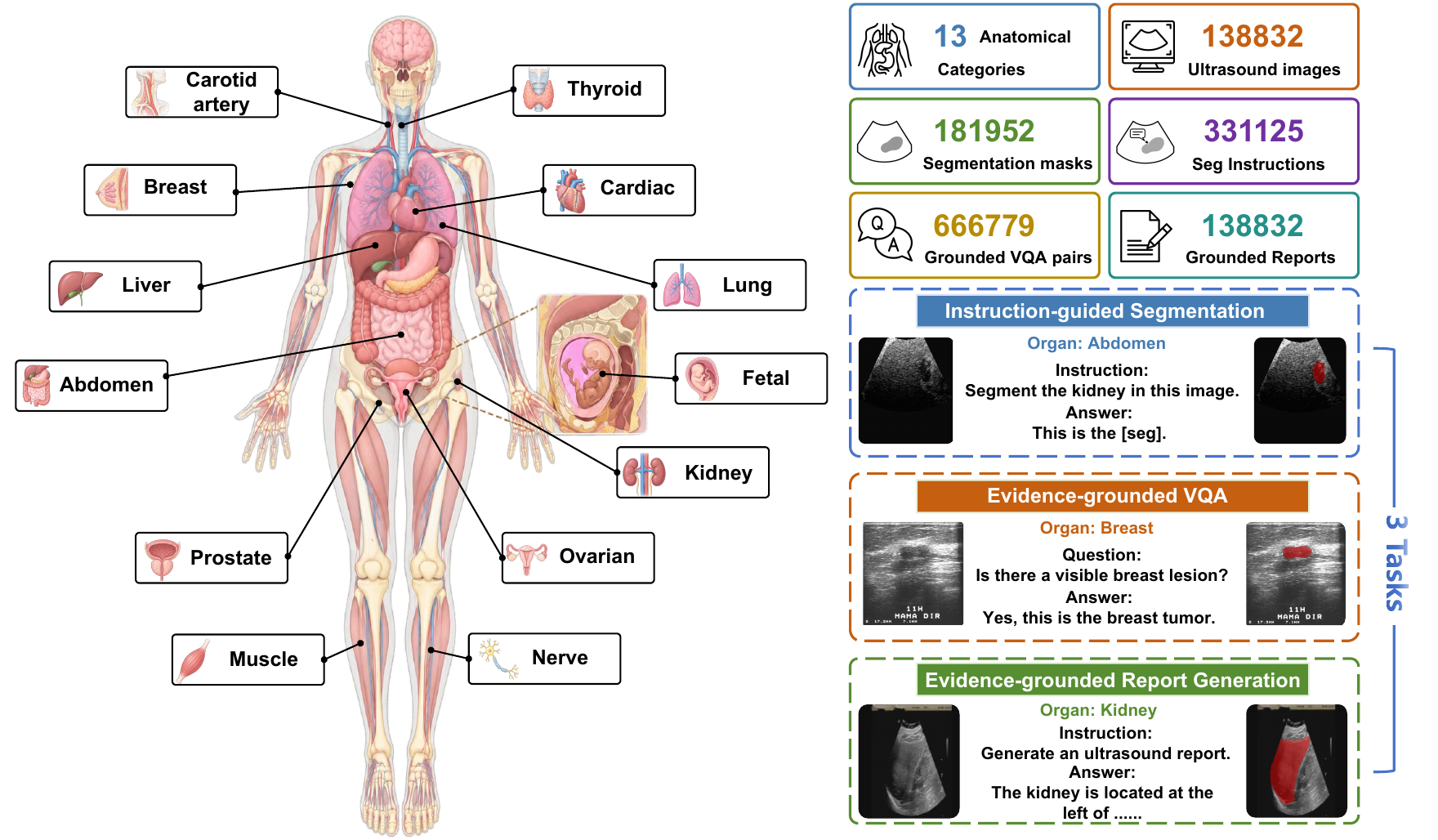}
    \caption{Overview of UltraG-Bench, including 13 anatomical categories, benchmark statistics, and representative examples of the three tasks: Instruction-guided Segmentation, Evidence-grounded VQA, and Evidence-grounded Report Generation.}
    \label{fig:introduction}
\end{figure}

Existing benchmarks often evaluate semantic understanding and pixel-level visual localization in isolation~\cite{lau2018dataset,liu2021slake,he2020pathvqa,hu2024omnimedvqa,li2026unim}: segmentation benchmarks primarily measure the overlap between predicted and ground-truth masks using metrics such as IoU and Dice, whereas VQA and report-generation benchmarks mainly assess the correctness of the generated answers or text. As a result, a model may produce the correct answer while relying on the wrong visual region, or generate fluent and semantically plausible descriptions without providing consistent pixel-level evidence for its claims. Recently, visual evidence-grounded datasets such as GEMeX~\cite{liu2025gemex} and MIMIC-ILS~\cite{choi2026instruction} have emerged, but such evaluation remains lacking in ultrasound, especially at the pixel level. 

To address these challenges, we introduce \textbf{UltraG-Bench} as shown in Figure~\ref{fig:introduction}, a large-scale multi-task benchmark for Pixel-level Evidence Grounding in ultrasound. UltraG-Bench annotates \textbf{40} public ultrasound segmentation datasets spanning \textbf{13} anatomical categories, including the abdomen, breast, heart, carotid artery, fetus, kidney, liver, lung, muscle, nerve, ovary, prostate, and thyroid, comprising \textbf{138,832} ultrasound images and \textbf{181,952} segmentation masks. Based on these pixel-level annotations, we construct three tasks from visual perception to semantic understanding: instruction-guided segmentation, which evaluates whether a model can localize the correct target from a natural-language instruction; evidence-grounded VQA, which assesses answer correctness and the accuracy of the corresponding visual evidence; evidence-grounded report generation, which examines whether a model can consistently associate different semantic statements with the appropriate pixel-level regions during report generation. During dataset construction, we explicitly preserve the correspondence between textual statements and the original segmentation masks, while restricting diagnostic, pathological, and treatment-related claims that are unsupported by the available pixel-level annotations, thereby ensuring that each grounded statement is backed by traceable visual evidence. Based on this benchmark, we systematically evaluate 15 state-of-the-art models, including closed-source general-purpose MLLMs, open-source general-purpose MLLMs, medical MLLMs, and pixel-level grounded medical MLLMs.

Furthermore, we observe a clear complementarity between high-level semantic understanding and fine-grained pixel localization in existing models: MLLMs excel at interpreting complex natural-language instructions and generating semantic responses, whereas specialized ultrasound segmentation models provide stronger pixel-level structural localization. This motivates a further question: Can more reliable Pixel-level Evidence Grounding be achieved through collaboration among pretrained models? To explore this, we construct \textbf{UltraG-Agent}, an ultrasound grounding agent that uses an open-source MLLM for instruction understanding, task planning, and response generation, while employing UltraSAM3~\cite{xu2026ultrasam3}, an ultrasound concept-driven segmentation model, as the visual module for target segmentation. Experiments show that UltraG-Agent achieves state-of-the-art performance on UltraG-Bench.

The main contributions are as follows:

\begin{itemize}
    \item \textbf{UltraG-Bench:} We annotate 40 public ultrasound segmentation datasets spanning 13 anatomical categories, comprising 138,832 ultrasound images and 181,952 segmentation masks. We construct three tasks, including instruction-guided segmentation, mask-grounded VQA, and mask-grounded report generation, to evaluate capabilities of existing models.
    \item \textbf{Comprehensive evaluation:} Under a unified evaluation framework, we benchmark 16 state-of-the-art models, including closed-source and open-source general MLLMs, medical MLLMs, pixel-level grounded medical MLLMs, and prompt-driven segmentation models.
    \item \textbf{UltraG-Agent:} We propose an ultrasound agent framework that combines the instruction understanding and task-planning capabilities of an open-source MLLM with the specialized ultrasound segmentation capability of UltraSAM3. UltraG-Agent achieves state-of-the-art performance among all evaluated models.
\end{itemize}

\section{Related Work}

\paragraph{Medical Evidence-Grounded Benchmarks.} Early medical VQA datasets, such as VQA-RAD, SLAKE, PathVQA, and OmniMedVQA, pair medical images with questions and answers~\cite{lau2018dataset,liu2021slake,he2020pathvqa,hu2024omnimedvqa}. They provide important benchmarks for medical visual understanding and question answering. As interpretability and visual evidence become more important, later datasets begin to link text with local image regions. MS-CXR, Chest ImaGenome, and PadChest-GR provide annotations that connect radiology findings, anatomical structures, and imaging attributes with corresponding bounding boxes in chest X-rays~\cite{boecking2022making,wu2021chest,de2025padchest}. These annotations enable image–text alignment to be evaluated at the region level. For VQA, GEMeX~\cite{liu2025gemex} introduces textual explanations and visual regions into Med-VQA. GEMeX-RMCoT~\cite{liu2025gemexcot} further links intermediate reasoning steps to relevant anatomical regions. MIMIC-ILS~\cite{choi2026instruction} extends referring image segmentation to chest X-rays, providing finer pixel-level evidence. However, existing benchmarks focus on radiological modalities, while ultrasound remains lacking. Moreover, most datasets focus on a single task. In contrast, UltraG-Bench builds a multi-task Pixel-level Evidence Grounding benchmark from 40 public ultrasound segmentation datasets. We have included a comparison with existing medical evidence-grounded benchmarks in Table~\ref{tab:dataset_comparison}.

\begin{table}[!ht]
\centering
\caption{
Comparison of UltraG-Bench with representative medical evidence-grounded datasets.
}
\label{tab:dataset_comparison}
\resizebox{\textwidth}{!}{
\begin{tabular}{lccccccccc}
\toprule
\textbf{Benchmark}
& \textbf{Modality}
& \textbf{Scale}
& \textbf{Spatial Evidence}
& \textbf{Region Grounding}
& \textbf{Instr.-Seg.}
& \textbf{Grounded VQA}
& \textbf{Grounded Report}
& \textbf{Coverage} \\
\midrule

MS-CXR
& CXR
& 1,162 pairs
& Bounding Box
& \cmark
& \xmark
& \xmark
& \xmark
& Chest \\

Chest ImaGenome
& CXR
& 242,072 images
& Bounding Box
& \cmark
& \xmark
& \xmark
& \xmark
& Chest \\

PadChest-GR
& CXR
& 4,555 studies
& Bounding Box
& \cmark
& \xmark
& \xmark
& \cmark
& Chest \\

GEMeX
& CXR
& 1.61M VQA pairs
& Bounding Box
& \cmark
& \xmark
& \cmark
& \xmark
& Chest \\

GEMeX-RMCoT
& CXR
& 202K instances
& Bounding Box
& \cmark
& \xmark
& \cmark
& \xmark
& Chest \\

MIMIC-ILS
& CXR
& 1.1M pairs
& Pixel Mask
& \cmark
& \cmark
& \xmark
& \xmark
& Chest \\

\midrule

\textbf{UltraG-Bench}
& \textbf{Ultrasound}
& \textbf{1.14M annotations}
& \textbf{Pixel Mask}
& \cmark
& \cmark
& \cmark
& \cmark
& \textbf{13 anatomies} \\

\bottomrule
\end{tabular}
}
\end{table}

\begin{figure}[!ht]
    \centering
    \includegraphics[width=\textwidth]{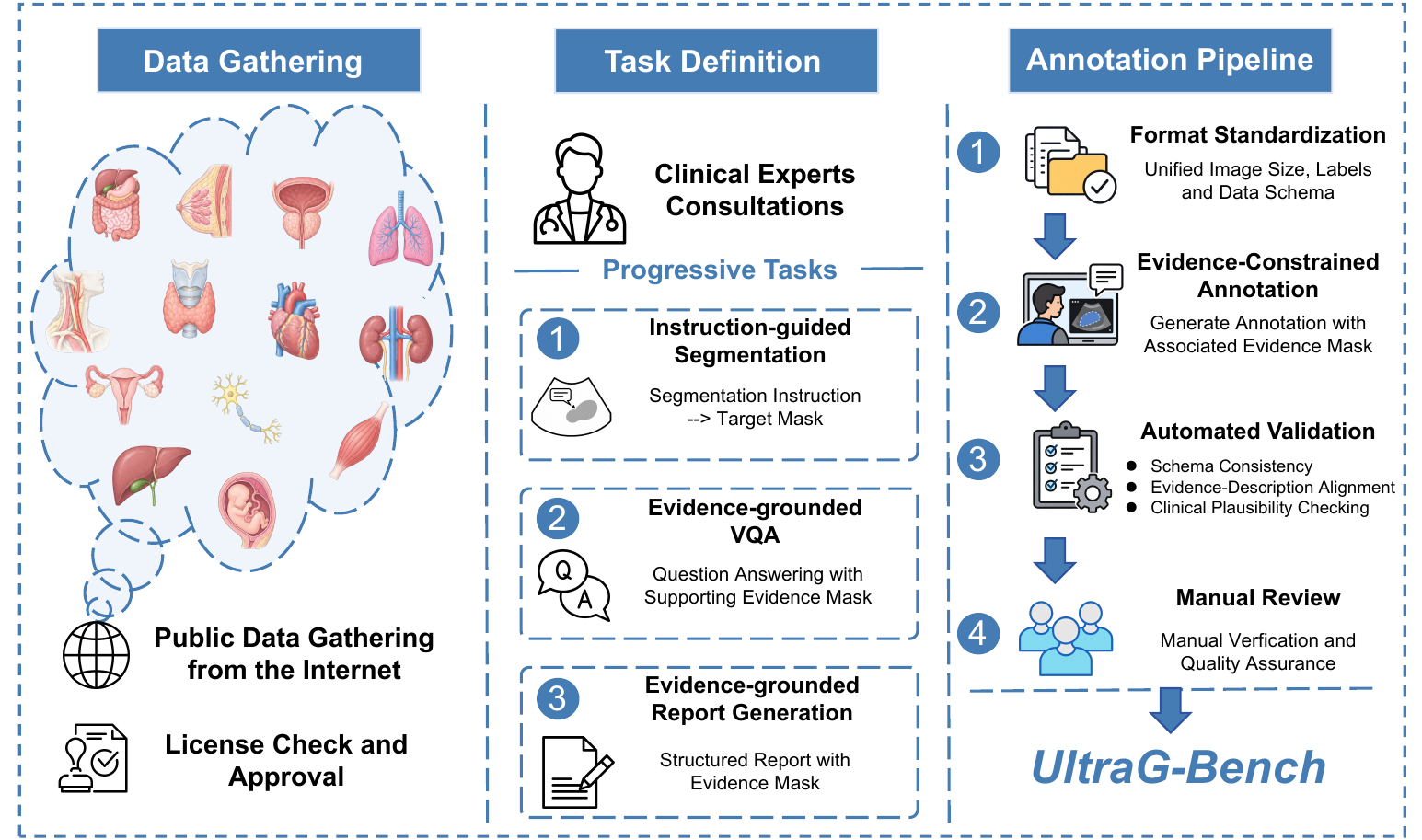}
    \caption{Construction pipeline of the UltraG-Bench, including public data collection, progressive task design, evidence-constrained annotation, automated validation, and expert review.}
    \label{fig:UltraG-Bench}
\end{figure}

\paragraph{Large Vision-Language Models for Medical Evidence Grounding.} Recently, both general-purpose and medical multimodal large language models have made rapid progress in medical image understanding. General MLLMs such as GPT~\cite{singh2025openai}, Claude~\cite{anthropic2025system}, Qwen-VL~\cite{bai2025qwen3}, InternVL~\cite{wang2025internvl3}, and MiniCPM-V~\cite{yao2024minicpm} show strong visual understanding, natural-language interaction, and open-ended reasoning. Medical MLLMs such as LLaVA-Med~\cite{li2023llava}, MedGemma~\cite{sellergren2025medgemma}, and Lingshu~\cite{xu2026lingshu} further improve medical image and clinical understanding through medical pretraining or instruction tuning. However, these models are often unclear whether their predictions are supported by the correct visual evidence. To provide fine-grained spatial outputs, another line of work endows MLLMs with segmentation capability. LISA~\cite{lai2024lisa} connects special tokens such as `[SEG]` to a segmentation decoder and enables reasoning segmentation from complex language instructions. Later models such as UniBiomed~\cite{wu2026universal} extend this idea to medical images and produce pixel-level masks from medical text instructions. However, their true capability for ultrasound Pixel-level Evidence Grounding remains underexplored. More recently, prompt-driven segmentation models such as SAM3~\cite{carion2026sam} have emerged. Inspired by this paradigm, we develop UltraG-Agent to evaluate agent-based grounding on our benchmark.

\section{UltraG-Bench}

\paragraph{Overview.}
We construct \textbf{UltraG-Bench}, a multi-task benchmark for evaluating \textbf{Pixel-level Evidence Grounding} in ultrasound images. UltraG-Bench is built upon 40 publicly available ultrasound image segmentation datasets, covering 13 anatomical categories, including the abdomen, breast, heart, carotid artery, fetus, kidney, liver, lung, muscle, nerve, ovary, prostate, and thyroid. Based on the pixel-level segmentation masks provided by the original datasets, we further construct three progressively challenging tasks that bridge visual perception and semantic understanding. The overall construction pipeline is illustrated in Figure~\ref{fig:UltraG-Bench}.

\subsection{Task Definition}

\paragraph{Instruction-guided Segmentation.}
Instruction-guided Segmentation evaluates whether a model can recognize and segment a target structure according to a natural-language instruction. Given an ultrasound image and a textual instruction describing the target object, such as \emph{``Segment the kidney in this image.''}, the model is required to understand the target concept specified by the instruction and output the corresponding pixel-level segmentation region. Each instruction is explicitly associated with one or more target masks from the original dataset. Therefore, this task jointly evaluates natural-language instruction understanding and fine-grained spatial localization.

\paragraph{Evidence-grounded VQA.}
Evidence-grounded VQA extends conventional medical visual question answering by additionally requiring pixel-level visual evidence that supports the generated answer. Given an ultrasound image and a related question, the model is required to generate a textual answer while simultaneously segmenting the region that supports the answer. Unlike conventional VQA benchmarks that evaluate only answer correctness, UltraG-Bench explicitly preserves the correspondence among the question, answer, and original segmentation mask. Consequently, even if a model produces a semantically correct answer, it is not considered to achieve reliable evidence grounding if the predicted visual evidence does not align with the ground-truth target region. This task therefore jointly evaluates semantic understanding and answer-evidence consistency.

\paragraph{Evidence-grounded Report Generation.}
Evidence-grounded Report Generation further extends pixel-level visual grounding to long-form text generation. Given an ultrasound image and a report-generation instruction, the model is required to generate a structured ultrasound description and associate image-supported semantic statements with their corresponding pixel-level evidence regions. Specifically, statements in the generated report are linked to target masks, establishing fine-grained \textbf{statement--evidence} correspondence. This task evaluates not only the semantic quality of the generated text, but also whether the model can maintain consistency between textual descriptions and pixel-level visual evidence throughout a longer generation process. Diagnostic, pathological, treatment-related, or other clinical inferences that cannot be supported by the available pixel-level annotations are excluded from grounded statements, thereby avoiding supervision based on untraceable visual evidence.

\subsection{Dataset Construction}

\subsubsection{Dataset Collection and Preprocessing}

We collect publicly available ultrasound image segmentation datasets. Before incorporating them into UltraG-Bench, we verify their public-use licenses and organize their images, segmentation annotations, and target-category information. The resulting collection covers 13 anatomical categories. Detailed descriptions of our benchmark, including data sources, target categories, and statistics, are provided in the Appendix~\ref{dataset}.

Since the source datasets differ substantially in image organization, mask representation, category naming, and annotation format, we first perform unified preprocessing. Specifically, segmentation annotations from different sources are converted into a COCO-style annotation format. Each image is assigned a unique identifier, and every segmentation mask is explicitly associated with its corresponding target-category name. The standardized representation contains unified image information, target categories, segmentation regions, and annotation identifiers, providing a consistent pixel-level evidence representation for the three downstream tasks. After standardization, UltraG-Bench contains 138,832 ultrasound images and 181,952 pixel-level segmentation masks. These standardized target-mask correspondences serve as the foundation for subsequent annotation.

\subsubsection{Evidence-constrained Annotation}

Based on the standardized segmentation data, we further construct annotations for the three tasks. Instead of allowing an MLLM to freely generate questions or reports, we adopt an Evidence-constrained Annotation strategy. For each image, the existing target-category names and their corresponding segmentation masks are first extracted and used as the factual basis available during annotation generation. The detailed annotation prompts are provided in the Appendix~\ref{sec:benchmark_construction}.

We use GPT-5.5~\cite{singh2025openai} to generate annotations. For each sample, the model is only allowed to construct segmentation instructions, VQA pairs, and grounded reports based on the provided target names and pixel-level evidence. For Instruction-guided Segmentation, we generate target segmentation instructions with diverse expressions while preserving their correspondence with the target masks. For example, \emph{``Segment the liver.''} is directly grounded by the liver mask.

\begin{figure}[!ht]
    \centering
    \includegraphics[width=\textwidth]{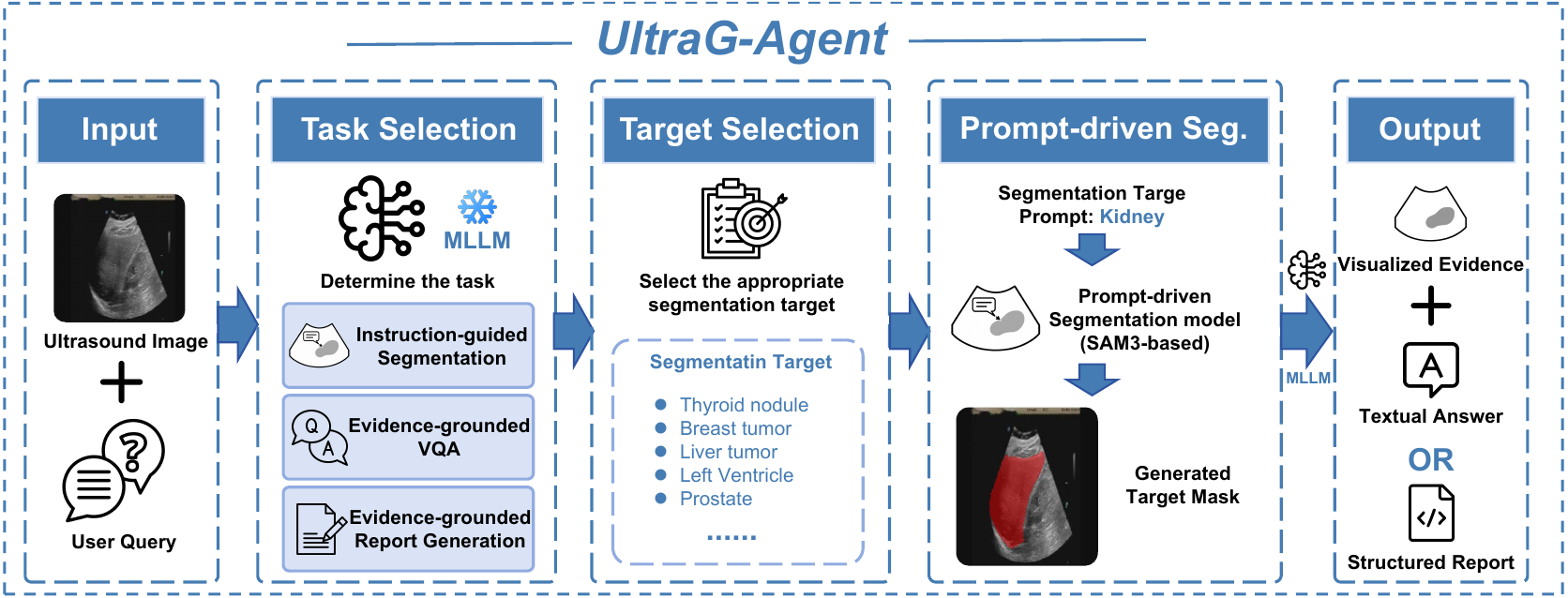}
    \caption{Overview of UltraG-Agent.}
    \label{fig:ultrag-agent}
\end{figure}

For \textbf{Evidence-grounded VQA}, we construct questions and answers related to visible targets, spatial locations, object counts, and other information supported by the available annotations. For example, given the question \emph{``Which abdominal structure occupies the upper-central region of the image?''}, the answer is \emph{``Liver.''}, and the corresponding liver mask is used as supporting visual evidence.

For \textbf{Evidence-grounded Report Generation}, we generate structured ultrasound descriptions based on the target categories and their pixel-level masks, focusing primarily on four types of information: presence, category, location, and size. For example, the model may generate \emph{``A visible kidney target is present in the middle-right region, with an approximate size of 77 $\times$ 45 pixels.''} The key information contained in the report is therefore directly supported by the ground-truth masks. 

The entire annotation process follows a strict evidence constraint. The annotation model is not allowed to introduce diagnoses, pathological attributes, treatment recommendations, or other medical conclusions that cannot be supported by the available pixel-level annotations. In other words, segmentation masks are primarily used to support visual facts such as target presence, spatial location, and extent, rather than serving as evidence for additional clinical attributes. In this way, the annotations are not generated independently of the segmentation labels, but remain traceably connected to the original pixel-level annotations, ensuring that every grounded prediction in UltraG-Bench can be traced back to explicit visual evidence.

\subsubsection{Dataset Validation}

To ensure annotation reliability, we adopt a two-stage validation pipeline consisting of automated validation and expert review. Automated checks assess data completeness, formatting consistency, image–annotation correspondence, and task-specific alignment between textual annotations and pixel-level evidence. Samples that fail these checks are removed or regenerated.

For manual review, two reviewers with training in ultrasound imaging independently assess 10\% samples from each anatomical structure, with representative coverage across categories and annotation types. They verify the consistency between annotations and their corresponding pixel-level evidence across all three tasks. Disagreements are jointly reviewed to reach consensus.

\subsection{Statistics}

\textbf{UltraG-Bench} contains \textbf{331,125} instruction-guided segmentation instructions, \textbf{666,779} evidence-grounded VQA pairs, and \textbf{138,832} evidence-grounded reports, forming a multi-level evaluation framework that ranges from target segmentation and visual question answering to report generation. Detailed statistics are provided in Appendix~\ref{dataset}.

\section{UltraG-Agent}

\textbf{SAM3} ~\cite{carion2026sam} is a prompt-driven segmentation model that supports open-vocabulary text prompts. It can perform segmentation according to text prompts, and further employs an MLLM-based agent to translate complex user instructions into executable segmentation targets. Inspired by this framework, we further propose \textbf{UltraG-Agent}, which combines the semantic understanding capability of a multimodal large language model with the fine-grained pixel-level localization capability of a prompt-driven ultrasound segmentation model to achieve more reliable Pixel-level Evidence Grounding. UltraG-Agent does not require additional end-to-end training. Instead, it consists of a pretrained MLLM without task-specific fine-tuning and UltraSAM3~\cite{xu2026ultrasam3}, a prompt-driven segmentation model adapted to the ultrasound domain. The overall framework is illustrated in Figure~\ref{fig:ultrag-agent}.

Given an ultrasound image and a user instruction, the MLLM first performs \textbf{task selection} to determine whether the current request corresponds to Instruction-guided Segmentation, Evidence-grounded VQA, or Evidence-grounded Report Generation. It then selects the \textbf{segmentation target} required for the task based on the image, the user instruction, and a predefined set of ultrasound target categories, and converts the selected target into a concise text prompt for segmentation. Depending on the task, the MLLM can select one or multiple targets and generate a corresponding UltraSAM3 prompt for each target.

After target selection, UltraSAM3 performs \textbf{prompt-driven segmentation} using the generated text prompts and produces pixel-level masks for the selected targets. For segmentation tasks, the predicted masks are directly returned as the output. For VQA and report generation, the system further extracts visual information from the predicted masks, including target presence, category, spatial location, and extent, and uses these mask-derived facts as reliable visual evidence to constrain the final text generation. UltraG-Agent ultimately outputs the visualized pixel-level evidence together with the corresponding textual answer or structured report, thereby combining the high-level semantic understanding capability of the MLLM with the fine-grained spatial localization capability of UltraSAM3. Detailed formulations and prompting strategies are provided in the Appendix~\ref{sec:ultrag_agent_details}.

\section{Experiment}

\subsection{Experimental Setting}
We evaluate \textbf{UltraG-Bench} on 14 representative MLLMs, including closed-source models, open-source general-purpose models, medical MLLMs, and pixel-level grounded medical MLLMs. To examine whether UltraG-Agent can extend prompt-driven segmentation models with VQA and report generation capabilities while preserving or improving their segmentation performance, we additionally evaluate BiomedParse~\cite{zhao2025foundation} and UltraSAM3~\cite{xu2026ultrasam3} as prompt-driven segmentation baselines. We further evaluate four UltraG-Agent variants with different MLLMs and prompt-driven segmentation models. All models are evaluated under a unified protocol across the three tasks of UltraG-Bench. Detailed baseline introduction and the prompts used for evaluation are given in Appendix~\ref{sec:baseline_intro} and Appendix~\ref{sec:evaluation_details}, respectively.

\subsection{Evaluation Metric}

\paragraph{Instruction-guided Segmentation.}
We use \textbf{Intersection over Union (IoU)} and \textbf{Dice coefficient (Dice)} to evaluate the overlap between the predicted masks and the ground-truth masks.

\paragraph{Evidence-grounded VQA.}
We report \textbf{Answer Accuracy (Acc.)} to evaluate the semantic correctness of the generated answers. Since a correct answer does not necessarily imply that the model relies on the correct visual evidence, we further introduce \textbf{Grounded Accuracy (G-Acc.)}. A prediction is counted as correct by G-Acc. only when both the textual answer is correct and the predicted evidence region is sufficiently consistent with the corresponding ground-truth mask. Specifically,
\begin{equation}
    \mathrm{G\text{-}Acc.}
    =
    \frac{1}{N}
    \sum_{i=1}^{N}
    \mathbb{I}
    \left(
    a_i=\hat{a}_i
    \land
    \mathrm{IoU}(\hat{M}_i,M_i)\geq\tau
    \right),
\end{equation}
where $\tau$ is the predefined grounding \textbf{threshold 0.5}.

\begin{table}[!ht]
\centering
\caption{
Experimental Results on UltraG-Bench. The best, second-best, and third-best results among all models are
highlighted in \rankone{green}, \ranktwo{blue}, and \rankthree{yellow}, respectively.
For UltraG-Agent, $I$, $Q$, and $L$ refer to InternVL3.5-8B, Qwen3-VL-8B, and Lingshu-7B, while $US3$ and $S3$ refer to UltraSAM3 and SAM3, respectively.
}
\label{tab:UltraG-Bench}

\setlength{\tabcolsep}{4pt}
\renewcommand{\arraystretch}{1.1}
\scriptsize

\resizebox{\linewidth}{!}{%
\begin{tabular}{lcccccc}
\toprule

\multirow{2}{*}{Model}
& \multicolumn{2}{c}{Instruction-guided Seg.}
& \multicolumn{2}{c}{Evidence-grounded VQA}
& \multicolumn{2}{c}{Evidence-grounded Report Gen.}
\\

\cmidrule(lr){2-3}
\cmidrule(lr){4-5}
\cmidrule(lr){6-7}

& IoU$\uparrow$
& Dice$\uparrow$
& Acc.$\uparrow$
& G-Acc.$\uparrow$
& ROUGE-L$\uparrow$
& SemAcc$\uparrow$
\\

\midrule

\multicolumn{7}{c}{\textit{Closed-source MLLMs}}\\
\midrule

Claude-4.5
& 0.1544
& 0.2180
& 0.3327
& 0.0214
& 0.1334
& 0.3281
\\

GPT-5.4
& 0.2653
& 0.3718
& 0.3631
& 0.0609
& 0.2264
& 0.3577
\\

\midrule

\multicolumn{7}{c}{\textit{Open-source General MLLMs}}\\
\midrule

Qwen3-VL-4B
& 0.1069
& 0.1537
& 0.2197
& 0.0137
& 0.0793
& 0.1899
\\

MiniCPM-V 2.6
& 0.0642
& 0.0854
& 0.0254
& 0.0000
& 0.1554
& 0.2592
\\

Qwen3-VL-8B
& 0.1061
& 0.1614
& 0.1938
& 0.0071
& 0.1274
& 0.2930
\\

Qwen3-VL-32B
& 0.1136
& 0.1595
& 0.3095
& 0.0113
& 0.2211
& 0.3025
\\

InternVL3.5-8B
& 0.0974
& 0.1462
& 0.3678
& 0.0081
& 0.1903
& 0.3113
\\

InternVL3.5-14B
& 0.1162
& 0.1755
& 0.3528
& 0.0176
& 0.2408
& 0.3426
\\

\midrule

\multicolumn{7}{c}{\textit{Open-source Medical MLLMs}}\\
\midrule

LLaVA-Med
& 0.1369
& 0.2006
& 0.2290
& 0.0000
& 0.1690
& 0.2822
\\

MedGemma-4B
& 0.1081
& 0.1702
& 0.3207
& 0.0025
& 0.1355
& 0.2938
\\

Lingshu-7B
& 0.0715
& 0.1114
& 0.3292
& 0.0000
& 0.1666
& 0.2975
\\

Lingshu-32B
& 0.0920
& 0.1413
& 0.2949
& 0.0046
& 0.1940
& 0.3000
\\

MedGemma-27B
& 0.1616
& 0.2424
& 0.3668
& 0.0056
& 0.2158
& 0.3436
\\

\midrule

\multicolumn{7}{c}{\textit{Pixel-level Grounded Medical MLLMs}}\\
\midrule

UniBiomed
& 0.1550
& 0.2068
& 0.0267
& 0.0002
& 0.1217
& 0.0567
\\

\midrule

\multicolumn{7}{c}{\textit{Prompt-driven Segmentation Models}}\\
\midrule

BiomedParse
& 0.2645
& 0.3303
& - -
& - -
& - -
& - -
\\

UltraSAM3
& \ranktwo{0.5478}
& \rankthree{0.6190}
& - -
& - -
& - -
& - -
\\

\midrule

\multicolumn{7}{c}{\textit{Ours}}\\
\midrule

UltraG-Agent$_{I+S3}$
& 0.0432
& 0.0483
& 0.0851
& 0.0000
& 0.1751
& 0.1568
\\

UltraG-Agent$_{L+US3}$
& 0.5292
& 0.6114
& \rankthree{0.5418}
& \rankthree{0.3351}
& \ranktwo{0.3805}
& \ranktwo{0.6225}
\\

UltraG-Agent$_{Q+US3}$
& \rankthree{0.5435}
& \ranktwo{0.6228}
& \ranktwo{0.5684}
& \ranktwo{0.4222}
& \rankthree{0.3300}
& \rankthree{0.5766}
\\

UltraG-Agent$_{I+US3}$
& \rankone{0.5564}
& \rankone{0.6384}
& \rankone{0.5970}
& \rankone{0.4531}
& \rankone{0.3971}
& \rankone{0.6687}
\\

\bottomrule
\end{tabular}%
}
\end{table}

\paragraph{Evidence-grounded Report Generation.}
We use \textbf{ROUGE-L} to measure the lexical similarity between the generated and reference reports. To further evaluate whether the generated report correctly captures the key visual evidence, we introduce \textbf{Semantic Accuracy (SemAcc)}. Specifically, we assess four semantic attributes: \textbf{presence}, which indicates whether the target is correctly identified as present or absent; \textbf{category}, which evaluates whether the target structure is correctly recognized; \textbf{location}, which measures whether the spatial position of the target is correctly described; and \textbf{size}, which evaluates whether the reported target extent is consistent with the ground-truth mask. These attributes correspond to binary indicators
$\mathrm{presence\_ok}$,
$\mathrm{category\_ok}$,
$\mathrm{location\_ok}$, and
$\mathrm{size\_ok}$.
The report-level semantic accuracy is defined as
\begin{equation}
    \mathrm{SemAcc}
    =
    \frac{1}{4}
    \left(
    \mathrm{presence\_ok}
    +
    \mathrm{category\_ok}
    +
    \mathrm{location\_ok}
    +
    \mathrm{size\_ok}
    \right),
\end{equation}
and the final score is averaged over all evaluated samples. Detailed definitions and evaluation of the four attributes are provided in the Appendix~\ref{sec:semantic_accuracy_details} and Appendix~\ref{sec:semacc_analysis}, respectively.

\subsection{Evaluation Results}

\subsubsection{Main Results}

\begin{figure}[!ht]
    \centering
    \includegraphics[width=\textwidth]{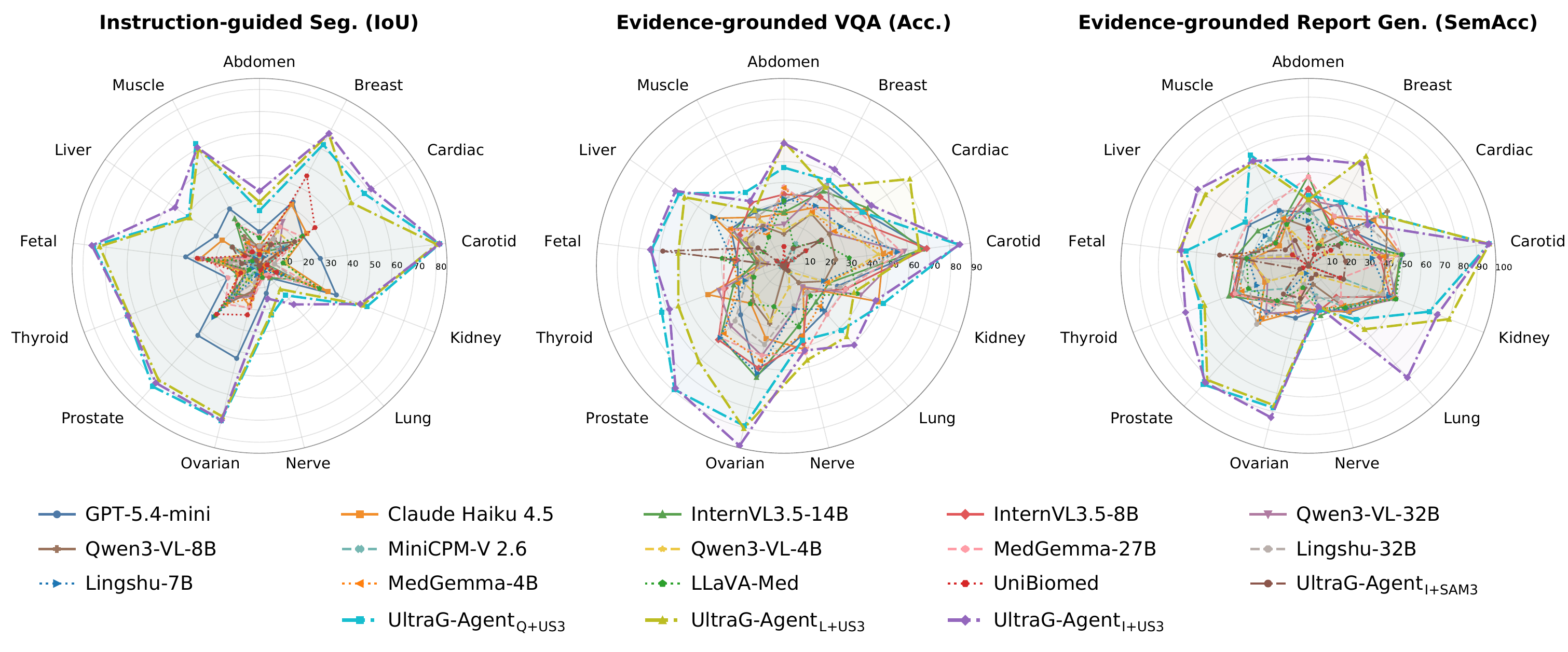}
    \caption{Fine-grained organ-level comparison across 13 anatomical categories for the three tasks in UltraG-Bench. UltraG-Agent variants with UltraSAM3 show consistently strong performance across diverse organs.}
    \label{fig:radar}
\end{figure}

Table~\ref{tab:UltraG-Bench} summarizes the overall performance of different models on the three tasks of UltraG-Bench. General-purpose MLLMs show relatively strong semantic understanding but limited pixel-level grounding. For example, InternVL3.5-8B and GPT-5.4 achieve VQA accuracies of 0.3678 and 0.3631, while their G-Acc. scores remain only 0.0081 and 0.0609, respectively. Medical MLLMs exhibit a similar pattern: models such as MedGemma-27B achieve competitive VQA and report-generation performance, but all obtain G-Acc. below 0.006 and relatively low segmentation IoU. In contrast, the pixel-level grounded medical model UniBiomed shows stronger segmentation capability but substantially weaker semantic performance. Prompt-driven segmentation models exhibit much stronger localization, with BiomedParse and UltraSAM3 achieving IoU/Dice scores of 0.2645/0.3303 and 0.5478/0.6190, respectively, but they cannot perform VQA or report generation.

In contrast, \textbf{UltraG-Agent} achieves substantial improvements across all three tasks, demonstrating the effectiveness of combining MLLM-based semantic understanding with ultrasound-specific pixel-level localization. Among all variants, UltraG-Agent$_{I+US3}$ achieves the best performance on all six metrics, outperforming the best baseline by 0.2911, 0.2666, 0.2292, 0.3922, 0.1563, and 0.3110 in IoU, Dice, Acc., G-Acc., ROUGE-L, and SemAcc, respectively. Notably, compared with UltraSAM3 alone, UltraG-Agent$_{I+US3}$ further improves IoU/Dice from 0.5478/0.6190 to 0.5564/0.6384, while additionally enabling VQA and report generation. Further comparison shows that UltraSAM3 is a key factor behind these gains: replacing it with the original SAM3 leads to substantial performance drops across segmentation, VQA, and report generation. Meanwhile, different MLLM backbones yield only modest differences and all maintain strong performance. Overall, these results indicate that reliable Pixel-level Evidence Grounding depends on both high-level semantic reasoning and specialized ultrasound pixel-level localization, with UltraSAM3 serving as a critical foundation.

\subsubsection{Fine-grained Organ-level Analysis}

Figure~\ref{fig:radar} presents the fine-grained performance of different models across 13 anatomical categories. Clear performance variations can be observed across organs, indicating that the difficulty of Pixel-level Evidence Grounding is jointly affected by anatomical structure and task type. In addition, both general-purpose and medical MLLMs exhibit substantial performance fluctuations across organs, particularly on Instruction-guided Segmentation, where their pixel-level localization ability is generally limited. Detailed case study and error analysis are provided in Appendix~\ref{case_study} and \ref{error_analysis}.

In contrast, UltraG-Agent variants with UltraSAM3 achieve broader radar coverage across most anatomical categories and show relatively consistent trends across different MLLM backbones. This indicates that the advantage of UltraG-Agent is not driven by only a few specific organs, but generalizes more consistently across diverse ultrasound anatomical structures.

\section{Conclusion}

In this work, we construct \textbf{UltraG-Bench}, a multi-task benchmark for evaluating Pixel-level Evidence Grounding in ultrasound. By annotating 40 public ultrasound segmentation datasets, UltraG-Bench provides unified evaluation across Instruction-guided Segmentation, Evidence-grounded VQA, and Evidence-grounded Report Generation. Our experiments reveal a clear gap between semantic understanding and pixel-level localization in existing MLLMs. To bridge this gap, we further propose \textbf{UltraG-Agent}. Extensive experiments show that UltraG-Agent consistently improves both semantic prediction and visual grounding performance.




\subsection*{Reproducibility statement}

To support reproducibility, detailed dataset information and baseline descriptions are provided in Appendix~\ref{dataset} and~\ref{sec:baseline_intro}. Benchmark construction details, including annotation prompts and quality validation, are described in Appendix~\ref{sec:benchmark_construction}. The implementation details of UltraG-Agent and the complete evaluation protocols are provided in Appendix~\ref{sec:ultrag_agent_details} and~\ref{sec:evaluation_details}, respectively. Detailed definitions and attribute-wise analyses of Semantic Accuracy are given in Appendix~\ref{sec:semantic_accuracy_details} and~\ref{sec:semacc_analysis}. Additional case studies and error analyses are provided in Appendix~\ref{case_study} and~\ref{error_analysis}. Our benchmark and associated code have been released anonymously at \url{https://anonymous.4open.science/r/UltraG-Bench-F908}.



\subsection*{The Use of Large Language Models}

We used ChatGPT only for language refinement and writing assistance. All generated contents were carefully reviewed and revised.

\bibliography{references}
\bibliographystyle{unsrt}

\newpage

\appendix
\section{Appendix}

\subsection{Datasets Details}
\label{dataset}

\paragraph{Source Dataset Statistics.} Table~\ref{tab:appendix_dataset_statistics} summarizes the 40 ultrasound segmentation datasets spanning 13 organs that serve as sources for UltraG-Bench. The selected source data comprises 138,832 images and 181,952 segmentation masks across the training and test splits.

\begin{table}[!ht]
\centering
\caption{Statistics of the 40 ultrasound datasets retained in the Grounded-US annotations.}
\label{tab:appendix_dataset_statistics}
\setlength{\tabcolsep}{4pt}
\renewcommand{\arraystretch}{1.08}
\resizebox{\textwidth}{!}{%
\begin{tabular}{llrrrr}
\toprule
\rowcolor{gray!15}
\textbf{Organ} & \textbf{Dataset} & \textbf{Train Images} & \textbf{Train Masks} & \textbf{Test Images} & \textbf{Test Masks} \\
\midrule
Abdomen & AbdomenUS~\cite{orlando_abdomenus_ussimandsegm} & 784 & 3,079 & 198 & 781 \\
Breast & BrEast~\cite{pawlowska2024curated} & 204 & 402 & 52 & 102 \\
Breast & BUID~\cite{ardakani2023open} & 185 & 370 & 47 & 94 \\
Breast & BUSI~\cite{al2020dataset} & 624 & 1,056 & 156 & 274 \\
Breast & BUS\_BRA~\cite{gomez2024bus} & 1,500 & 3,000 & 375 & 750 \\
Breast & BUS\_DatasetB~\cite{yap2017automated} & 130 & 260 & 33 & 66 \\
Breast & BUS\_UC~\cite{iqbal2024memory} & 648 & 1,294 & 163 & 326 \\
Breast & BUS\_UCLM~\cite{vallez2025bus} & 546 & 408 & 137 & 120 \\
Breast & S1~\cite{guo2021segmentation} & 192 & 388 & 9 & 18 \\
\addlinespace[2pt]
Cardiac & CAMUS~\cite{leclerc2019deep} & 1,600 & 4,800 & 400 & 1,200 \\
Cardiac & CardiacUDC~\cite{yang2023graphecho} & 1,717 & 5,892 & 533 & 1,754 \\
Cardiac & EchoCP~\cite{wang2021echocp} & 434 & 1,696 & 132 & 521 \\
Cardiac & EchoNet\_Dynamic~\cite{ouyang2020video} & 17,476 & 17,476 & 2,550 & 2,550 \\
Cardiac & EchoNet\_Pediatric~\cite{reddy2023video} & 12,071 & 12,071 & 3,300 & 3,300 \\
Cardiac & Unity~\cite{unityimaging2024echocardiography} & 13,136 & 5,675 & 3,283 & 1,367 \\
\addlinespace[2pt]
Carotid artery & CCA~\cite{bi2023mi} & 2,307 & 2,195 & 540 & 540 \\
\addlinespace[2pt]
Fetal & ACOUSLIC~\cite{sappia2025acouslic} & 5,327 & 5,327 & 1,293 & 1,293 \\
Fetal & FASS~\cite{da2023fetal} & 1,270 & 10,012 & 318 & 2,506 \\
Fetal & Fast\_UNet~\cite{ashkani2022fast} & 1,128 & 1,128 & 283 & 283 \\
Fetal & HC~\cite{van2018automated} & 799 & 799 & 200 & 200 \\
Fetal & fh\_ps~\cite{jieyun_2024_10829116} & 3,200 & 6,400 & 800 & 1,600 \\
Fetal & focus~\cite{songxiong2025focus} & 250 & 500 & 50 & 100 \\
\addlinespace[2pt]
Kidney & KidneyUS~\cite{singla2023open} & 388 & 388 & 98 & 98 \\
Kidney & Ultrasound\_Normal\_Kidney~\cite{jeevaws2025ultrasoundnormalkidney} & 864 & 1,728 & 216 & 432 \\
\addlinespace[2pt]
Liver & Annotated\_Ultrasound\_Liver~\cite{xu_yiming_2022_7272660} & 588 & 510 & 147 & 125 \\
Lung & LUSS~\cite{mclaughlan2024lung} & 451 & 2,977 & 113 & 775 \\
\addlinespace[2pt]
Muscle & FALLMUD~\cite{cunningham2020fallmud} & 650 & 655 & 163 & 326 \\
Muscle & LUMINOUS~\cite{belasso2020luminous} & 269 & 302 & 68 & 78 \\
Muscle & STMUS\_NDA~\cite{marzola2021deep} & 6,529 & 6,530 & 1,632 & 1,632 \\
\addlinespace[2pt]
Nerve & UPBD~\cite{ding2022mallesnet} & 764 & 3,441 & 191 & 935 \\
\addlinespace[2pt]
Ovarian & OTU\_2d~\cite{zhao2022mmotu} & 1,175 & 1,191 & 294 & 298 \\
Ovarian & OTU\_3d~\cite{zhao2022mmotu} & 136 & 146 & 34 & 40 \\
\addlinespace[2pt]
Prostate & MicroSeg~\cite{shao2024micro} & 2,078 & 2,085 & 542 & 542 \\
Prostate & RegPro~\cite{baum2023mr} & 3,516 & 3,786 & 367 & 396 \\
\addlinespace[2pt]
Thyroid & DDTI~\cite{pedraza2015open} & 509 & 510 & 128 & 128 \\
Thyroid & KFGNet~\cite{wang2022key} & 131 & 161 & 33 & 37 \\
Thyroid & Segthy~\cite{kronke2022tracked} & 9,938 & 19,187 & 1,952 & 3,482 \\
Thyroid & TG3K~\cite{gong2023thyroid} & 2,868 & 3,056 & 717 & 772 \\
Thyroid & TN3K~\cite{gong2021multi} & 2,794 & 3,063 & 699 & 757 \\
Thyroid & Thyroid\_US\_Cineclip~\cite{stanford_aimi_thyroid_ultrasound_cine_clip_2026} & 13,565 & 13,565 & 3,845 & 3,845 \\
\midrule
\textbf{Total} & -- & \textbf{112,741} & \textbf{147,509} & \textbf{26,091} & \textbf{34,443} \\
\bottomrule
\end{tabular}%
}
\end{table}

\begin{figure}[!ht]
    \centering
    \includegraphics[width=\textwidth]{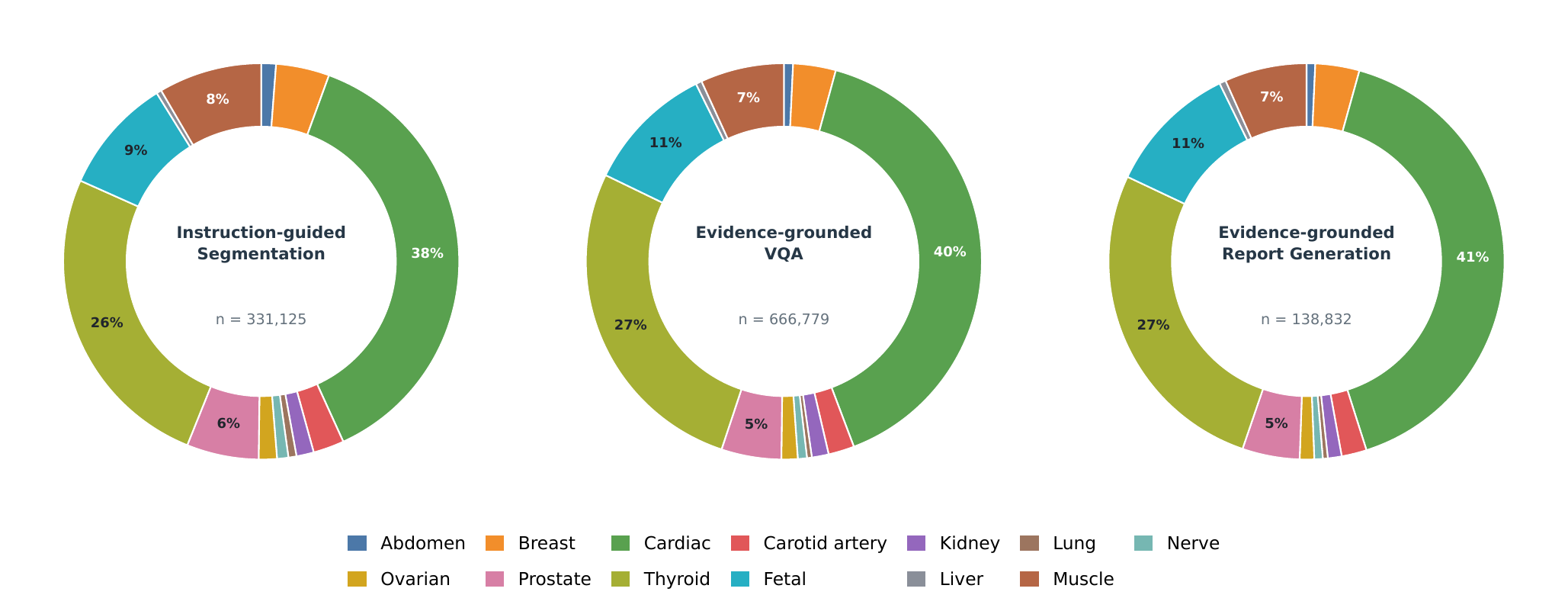}
    \caption{Anatomical-category distributions of the three tasks in UltraG-Bench across 13 anatomical categories.}
    \label{fig:task_distribution}
\end{figure}

\paragraph{Task-wise Anatomical Distribution.}
Figure~\ref{fig:task_distribution} shows the anatomical-category distributions of the three tasks in UltraG-Bench. Overall, the three tasks exhibit similar distributions across the 13 anatomical categories. Cardiac and thyroid samples constitute the largest proportions, accounting for 38\%--41\% and 26\%--27\% of the annotations, respectively, followed by fetal, muscle, and prostate categories. The remaining anatomical categories occupy smaller proportions. Despite differences in the total number of annotations across tasks, their anatomical compositions remain broadly consistent, providing comparable category coverage for evaluating pixel-level evidence grounding from segmentation to VQA and report generation.

\subsection{Baseline Introduction}
\label{sec:baseline_intro}

We evaluate a diverse set of baseline models covering closed-source general-purpose MLLMs, open-source general-purpose MLLMs, medical MLLMs, pixel-level grounded medical MLLMs, and prompt-driven segmentation models. The evaluated baselines are briefly introduced below.

\paragraph{Claude-4.5.}
Claude-4.5~\cite{anthropic2025system} is a proprietary multimodal large language model developed by Anthropic, with strong general-purpose capabilities in visual understanding and multimodal reasoning. We include it as a representative closed-source MLLM for evaluating semantic understanding and visual grounding in ultrasound images.

\paragraph{GPT-5.4.}
GPT-5.4~\cite{singh2025openai} is a proprietary frontier model developed by OpenAI with advanced multimodal understanding and reasoning capabilities. We evaluate GPT-5.4 as a strong closed-source general-purpose baseline across all three tasks of UltraG-Bench.

\paragraph{Qwen3-VL Series.}
Qwen3-VL~\cite{bai2025qwen3} is an open-source vision-language model family designed for general multimodal perception and reasoning, supporting diverse visual inputs and fine-grained visual understanding. We evaluate the 4B, 8B, and 32B variants to investigate the effect of model scale on ultrasound Pixel-level Evidence Grounding.

\paragraph{MiniCPM-V 2.6.}
MiniCPM-V 2.6~\cite{yao2024minicpm} is a compact 8B multimodal model designed for efficient image, multi-image, and video understanding. Despite its relatively small model size, it provides strong general-purpose visual-language capabilities and serves as an efficient open-source MLLM baseline.

\paragraph{InternVL3.5 Series.}
InternVL3.5\cite{wang2025internvl3} is an open-source multimodal model family that improves visual-language understanding, reasoning, and inference efficiency through enhanced multimodal training and reinforcement learning. We evaluate its 8B and 14B variants as representative general-purpose MLLMs with strong multimodal reasoning capabilities.

\paragraph{LLaVA-Med.}
LLaVA-Med~\cite{li2023llava} adapts a general-purpose vision-language assistant to the biomedical domain using large-scale biomedical image-text data and instruction tuning. It is designed for open-ended conversational understanding of biomedical images and serves as a representative early medical MLLM.

\paragraph{MedGemma Series.}
MedGemma~\cite{sellergren2025medgemma} is a family of medical vision-language foundation models built upon Gemma and specifically adapted for medical image and text understanding. We evaluate the 4B and 27B variants to assess whether medical-domain adaptation and increased model capacity improve ultrasound evidence grounding.

\paragraph{Lingshu Series.}
Lingshu~\cite{xu2026lingshu} is a generalist medical MLLM designed for unified multimodal medical understanding and reasoning through large-scale medical data curation and multi-stage training. We evaluate the 7B and 32B variants, which support a broad range of medical imaging modalities and tasks including medical VQA and report generation.

\paragraph{UniBiomed.}
UniBiomed~\cite{wu2026universal} is a grounded biomedical vision-language foundation model that jointly performs semantic interpretation and pixel-level segmentation of biomedical targets. Unlike conventional medical MLLMs that primarily generate textual responses, UniBiomed explicitly connects generated medical findings with localized visual regions, making it particularly relevant to Pixel-level Evidence Grounding.

\paragraph{BiomedParse.}
BiomedParse~\cite{zhao2025foundation} is a biomedical image parsing foundation model that unifies segmentation, detection, and recognition across diverse biomedical imaging modalities. It supports text-prompted segmentation of biomedical objects and is therefore included as a representative prompt-driven biomedical segmentation model.

\paragraph{UltraSAM3.}
UltraSAM3~\cite{xu2026ultrasam3} is a concept-driven foundation model specifically developed for universal ultrasound image segmentation. By adapting SAM3 to ultrasound-specific image--mask--concept triplets, it enables text-based target specification and provides strong prompt-driven pixel-level localization across diverse ultrasound anatomical structures.

\subsection{Benchmark Construction Details}
\label{sec:benchmark_construction}

\paragraph{Annotation Details.}
For each ultrasound image, we first extract evidence facts from the original COCO-format segmentation annotations, including target presence, category, mask identity, coarse image location, pixel extent, mask area ratio, boundary information, and the number of distinct targets. These mask-derived facts serve as the only visual evidence available during annotation. Based on them, we construct three types of annotations: instruction-guided segmentation, evidence-grounded VQA, and evidence-grounded report generation. The annotation process follows strict evidence constraints: generated text is restricted to information directly supported by the supplied labels and mask geometry, while unsupported diagnostic, pathological, treatment-related, or metadata-related statements are prohibited. After generation, all annotations are automatically normalized and validated to ensure valid evidence references and task-format consistency. Segmentation instructions and grounded reports are deterministically constructed from mask-derived facts, while VQA pairs are further filtered according to clinical relevance and evidence validity; invalid outputs are replaced by rule-based fallback annotations. The same annotation pipeline is applied to all 13 anatomical categories by replacing the organ-specific target with \texttt{\{organ\}}.

\begin{promptbox}
\small
\textbf{Prompt}

Create a grounded multi-task ultrasound sample for \texttt{\{organ\}}.
Return valid JSON only.

Use only the supplied facts. The \texttt{\{organ\}} label identifies the
segmentation target. Do not introduce any clinical attribute that is not
explicitly supported by the supplied labels or mask geometry.

Do not invent diagnoses, pathological findings, prognosis, treatment,
recommendations, or other unsupported clinical conclusions.

Never ask about or mention file names, case numbers, slice numbers,
image IDs, dataset/source names, split names, paths, dates, annotations,
mask IDs, or metadata in user-facing text.

Every VQA question must be directly supported by the available
\texttt{\{organ\}} evidence, such as:
\begin{itemize}
    \item target presence or visibility;
    \item image-region location;
    \item pixel extent or size;
    \item area ratio;
    \item boundary availability;
    \item number of visible targets.
\end{itemize}

Use \texttt{[SEG]} only when \texttt{target\_mask\_ids} is non-empty.
Mask IDs may appear only in ID fields.

The supplied masks support target presence, approximate image location,
pixel-level size and extent, target count, and boundary information.

For questions about a visible \texttt{\{organ\}} target, set
\texttt{requires\_grounding=true} and
\texttt{evidence\_source=mask\_geometry}.

If no target mask is available, generate only absence or
not-applicable questions with
\texttt{requires\_grounding=false}.

Return exactly the following top-level keys:

\texttt{segmentation\_tasks}: 2--4 objects with
\texttt{task\_id}, \texttt{instruction}, \texttt{answer},
\texttt{target\_mask\_ids}, and \texttt{answerability}.

\texttt{grounded\_vqa}: 4--6 objects with
\texttt{qa\_id}, \texttt{question}, \texttt{answer},
\texttt{answer\_type}, \texttt{requires\_grounding},
\texttt{evidence\_mask\_ids}, \texttt{reasoning\_type},
and \texttt{evidence\_source}.

\texttt{grounded\_report}: an object with a report prompt and
5--7 sentence objects containing
\texttt{section}, \texttt{text}, \texttt{evidence\_mask\_ids},
and \texttt{finding\_type}.

Facts:

\texttt{\{mask-derived evidence facts\}}
\end{promptbox}

\paragraph{Automated Quality Validation.}
We perform rule-based automatic quality validation for every generated sample before manual review. The validator checks five aspects. First, masks must have unique identifiers, valid anatomical categories, positive geometry and area, valid area ratios, coordinates within image boundaries, and no duplicated geometry. Second, each sample must contain 2--4 instruction-guided segmentation annotations, where every referenced target mask exists and the \texttt{[SEG]} token is used if and only if a valid mask is referenced. Third, each sample must contain 4--6 evidence-grounded VQA pairs with non-empty question--answer content, valid evidence masks, and consistent \texttt{requires\_grounding} flags. Fourth, grounded reports must contain 5--7 sentences with at least \textit{Findings} and \textit{Impression}, and all statements about target presence, category, location, extent, and count must agree with the referenced masks. Finally, we reject unsupported content beyond the available pixel-level evidence, including unannotated anatomical attributes, unsupported laterality or physical measurements, diagnostic or pathological claims, treatment recommendations, and metadata-related statements. Organ-specific constraints are further defined for each \texttt{\{organ\}}. Issues are categorized as critical, major, minor, or informational, with penalties of 25, 10, 3, and 0 points from an initial score of 100; samples containing any critical or major issue are flagged as failing automatic validation.

\subsection{UltraG-Agent Implementation Details}
\label{sec:ultrag_agent_details}

UltraG-Agent follows a modular inference pipeline that separates high-level
instruction understanding from pixel-level visual grounding. Given an ultrasound
image $I$, a user instruction $q$, and a predefined target-category set
$\mathcal{C}$, an MLLM first acts as a task planner to identify the task type and
the anatomical targets required to answer the instruction:
\begin{equation}
    \mathcal{P}
    =
    \mathcal{M}_{\mathrm{plan}}(I,q,\mathcal{C})
    =
    \left(
    \tau,
    \{(c_k,p_k)\}_{k=1}^{K}
    \right),
\end{equation}
where
$\tau \in
\{\text{Segmentation},\text{VQA},\text{Report}\}$
denotes the inferred task,
$c_k \in \mathcal{C}$ is the selected target category, and $p_k$ is a concise
text prompt generated for UltraSAM3. Multiple targets are selected only when
required by the instruction.

For each planned target, UltraSAM3 independently performs prompt-driven
segmentation:
\begin{equation}
    \{(m_{k,j},s_{k,j})\}_{j=1}^{N_k}
    =
    \mathcal{S}_{\mathrm{US3}}(I,p_k),
\end{equation}
where $m_{k,j}$ and $s_{k,j}$ denote the $j$-th candidate mask and its confidence
score, respectively. After filtering invalid candidates, the highest-scoring
mask is selected as
\begin{equation}
    \hat{m}_k
    =
    \arg\max_{m_{k,j}} s_{k,j}.
\end{equation}
We then extract mask-derived evidence
\begin{equation}
    g_k = \mathcal{G}(\hat{m}_k)
    =
    \{b_k,\, l_k,\, a_k,\, r_k,\, d_k^{\mathrm{long}},
    \,d_k^{\mathrm{short}}\},
\end{equation}
including the bounding box $b_k$, image-region location $l_k$, pixel area
$a_k$, area ratio $r_k$, and long- and short-axis pixel extents.
These quantities form the authoritative visual evidence used by the subsequent
response generation stage.

The final response is conditioned on the original instruction, the task plan,
and the mask-derived evidence:
\begin{equation}
    y =
    \mathcal{M}_{\mathrm{resp}}
    \left(
    I,q,\mathcal{P},
    \{(c_k,g_k)\}_{k=1}^{K}
    \right).
\end{equation}
For instruction-guided segmentation, the output consists of the textual
response and the corresponding predicted mask. For evidence-grounded VQA, the
answer is paired with the union of the masks referenced by the response:
\begin{equation}
    M_{\mathrm{VQA}}
    =
    \bigcup_{k\in\mathcal{T}_{a}}\hat{m}_k,
\end{equation}
where $\mathcal{T}_{a}$ denotes the targets supporting the answer.
For evidence-grounded report generation, each report statement $y_i$ is
associated with its own evidence mask,
\begin{equation}
    M_i =
    \bigcup_{k\in\mathcal{T}_{i}}\hat{m}_k,
\end{equation}
thereby preserving fine-grained statement--evidence correspondence throughout
the generated report. Mask-derived facts are treated as authoritative during
response generation, preventing the MLLM from modifying target presence,
location, pixel extent, or area measurements inferred from UltraSAM3.

\subsection{Evaluation Details}
\label{sec:evaluation_details}

\paragraph{Prompt-driven Segmentation Models Evaluation Details.}
For prompt-driven segmentation models, the ultrasound image and its segmentation instruction are directly provided as input. The predicted mask is evaluated against the ground-truth segmentation using IoU and Dice, following the same protocol as Instruction-guided Segmentation.

\paragraph{UltraG-Agent Evaluation Details.} The same evaluation samples, reference annotations, mask decoding procedure, and metric implementation are used for UltraG-Agent to ensure direct comparability with the MLLM baselines.

\paragraph{MLLM Evaluation Details.} We evaluate all MLLMs under a unified zero-shot protocol across the three tasks of UltraG-Bench. 
For each sample, the model receives the original ultrasound image together with a task-specific textual prompt and is required to return a structured JSON response. 
For spatial predictions, models may output either a polygon or a bounding box in the original-image pixel coordinate system; these predictions are rasterized into binary masks and compared with the corresponding ground-truth segmentation masks. 
All API-based models are evaluated with temperature $0$ to reduce sampling variance, and the same task prompts and output constraints are used across different models. 
No ground-truth mask coordinates or reference answers are provided to the model during inference.
For report generation, only the required finding-type structure is specified to standardize the output format, while the reference report content and evidence masks remain hidden.

For Instruction-guided Segmentation, we report IoU and Dice between the predicted and ground-truth masks. 
For Evidence-grounded VQA, Answer Accuracy measures semantic correctness, while Grounded Accuracy further requires the answer to be correct and the predicted evidence region to achieve an IoU above $0.5$ with the ground-truth evidence mask. 
For Evidence-grounded Report Generation, ROUGE-L evaluates textual similarity, while Semantic Accuracy measures the correctness of four evidence-supported attributes: target presence, category, location, and size. 
Location is evaluated using the predefined $3\times3$ image-region grid, and predicted pixel extent is considered correct when both dimensions are within a 20\% tolerance of the reference values.

\paragraph{Evaluation Prompts.}
The following prompts are used consistently for all evaluated MLLMs.
Here, \texttt{\{organ\}}, \texttt{\{W\}}, and \texttt{\{H\}} denote the
anatomical category and the original image width and height, respectively.

\begin{promptbox}
\small
\textbf{Instruction-guided Segmentation Prompt}

You are given a clinical ultrasound image and a segmentation instruction.

Return valid JSON only with keys: \texttt{answer}, \texttt{predicted\_mask}.

\texttt{predicted\_mask} must be
\texttt{\{"type":"polygon","points":[[x,y],...]\}},
\texttt{\{"type":"bbox","bbox\_xywh":[x,y,w,h]\}}, or \texttt{null}.

Use original-image pixel coordinates.
Include \texttt{[SEG]} in \texttt{answer} if a target is present.

Do not use annotation, annotated, dataset, mask id, or metadata.

\textbf{Input:}
\begin{verbatim}
{
  "image_width": {W},
  "image_height": {H},
  "instruction": "{instruction}"
}
\end{verbatim}
\end{promptbox}

\begin{promptbox}
\small
\textbf{Evidence-grounded VQA Prompt}

You are given a clinical ultrasound image and a question.

Return valid JSON only with keys: \texttt{answer}, \texttt{predicted\_mask}.

If the answer depends on a visible target, organ, or structure, provide
\texttt{predicted\_mask} as a polygon or bbox in original-image pixel
coordinates; otherwise use \texttt{null}.

Do not use annotation, annotated, dataset, mask id, or metadata.

\textbf{Input:}
\begin{verbatim}
{
  "organ": "{organ}",
  "image_width": {W},
  "image_height": {H},
  "question": "{question}"
}
\end{verbatim}
\end{promptbox}

\begin{promptbox}
\small
\textbf{Evidence-grounded Report Generation Prompt}

You are given a clinical ultrasound image.

Return valid JSON only with key \texttt{report}.

\texttt{report} must be a list of sentence objects with
\texttt{section}, \texttt{text}, \texttt{finding\_type}, and
\texttt{predicted\_mask}.

For spatial \texttt{finding\_type} values, \texttt{predicted\_mask} must
localize the supporting target, organ, or structure as a polygon or bbox
in original-image pixel coordinates.

Use \texttt{null} predicted mask for non-spatial statements.
Use the provided finding-type structure when applicable.

Use concise benchmark-style wording.
Do not invent ultrasound descriptors if uncertain.
Do not invent diagnostic grades, pathology classes, or treatment
recommendations, and do not use cm unless a scale marker is visible.
If estimating size, use pixel coordinates or pixel extent only.
Do not use annotation, annotated, dataset, mask id, or metadata.

\textbf{Input:}
\begin{verbatim}
{
  "organ": "{organ}",
  "image_width": {W},
  "image_height": {H},
  "prompt": "{report_instruction}"
}
\end{verbatim}
\end{promptbox}

\subsection{Detailed Definition of Semantic Accuracy}
\label{sec:semantic_accuracy_details}

To complement lexical similarity metrics, we evaluate whether a generated
report preserves the key visual evidence represented by the reference
annotation. Specifically, Semantic Accuracy (\textbf{SemAcc}) consists of four
binary attributes: target presence, anatomical category, spatial location, and
target size. Let $R$ and $\hat{R}$ denote the reference and generated reports,
respectively. Each attribute produces a binary indicator in $\{0,1\}$, as
defined below.

\paragraph{Target Presence.}
The presence attribute evaluates whether the generated report correctly
describes the target as present or absent. We extract a binary target-presence
state from both the reference and generated reports:
\begin{equation}
    p(R),\,p(\hat{R}) \in \{0,1\},
\end{equation}
where $1$ denotes that the target is present and $0$ denotes that the target is
absent. Presence correctness is defined as
\begin{equation}
    \mathrm{presence\_ok}
    =
    \mathbb{I}\left[p(\hat{R}) = p(R)\right],
\end{equation}
where $\mathbb{I}[\cdot]$ is the indicator function. In practice, explicit
absence expressions (e.g., ``no target'', ``target is absent'', or
``target is not visible'') are interpreted as negative evidence, whereas
explicit mentions of the target structure indicate target presence.

\paragraph{Anatomical Category.}
The category attribute measures whether the generated report correctly
identifies the anatomical target specified by the reference annotation.
Let $\mathcal{C}(R)$ and $\mathcal{C}(\hat{R})$ denote the sets of target
categories mentioned in the reference and generated reports, respectively.
We define
\begin{equation}
    \mathrm{category\_ok}
    =
    \mathbb{I}
    \left[
        \mathcal{C}(R)
        \subseteq
        \mathcal{C}(\hat{R})
    \right].
\end{equation}
Thus, all target categories required by the reference report must be correctly
identified in the generated report.

\paragraph{Spatial Location.}
We represent target location using a $3\times3$ spatial grid over the original
image. Given an image of width $W$ and height $H$, and the center of the
reference target mask $(x_c,y_c)$, its horizontal position is defined as
\begin{equation}
    h(x_c)=
    \begin{cases}
        \text{left}, & x_c < W/3,\\
        \text{central}, & W/3 \leq x_c < 2W/3,\\
        \text{right}, & x_c \geq 2W/3,
    \end{cases}
\end{equation}
and its vertical position is
\begin{equation}
    v(y_c)=
    \begin{cases}
        \text{upper}, & y_c < H/3,\\
        \text{middle}, & H/3 \leq y_c < 2H/3,\\
        \text{lower}, & y_c \geq 2H/3.
    \end{cases}
\end{equation}
The resulting location label is
\begin{equation}
    \ell=(v(y_c),h(x_c)),
\end{equation}
corresponding to one of nine regions such as
\textit{upper-left}, \textit{middle-central}, or \textit{lower-right}.
Let $\mathcal{L}(R)$ and $\mathcal{L}(\hat{R})$ denote the spatial-location
labels extracted from the reference and generated reports. Location
correctness is defined as
\begin{equation}
    \mathrm{location\_ok}
    =
    \mathbb{I}
    \left[
        \mathcal{L}(\hat{R})
        =
        \mathcal{L}(R)
    \right].
\end{equation}

\paragraph{Target Size.}
The size attribute evaluates whether the pixel-level extent reported by the
model is consistent with that derived from the ground-truth mask. Let
$(d_1,d_2)$ denote the two reference pixel extents and
$(\hat{d}_1,\hat{d}_2)$ denote the corresponding values extracted from the
generated report. Since the ordering of the long and short axes may vary in
natural-language generation, we consider both direct and reversed matching.
For each reference dimension $d$, the allowable tolerance is
\begin{equation}
    \delta(d)=\max(1,\;0.2d),
\end{equation}
i.e., either one pixel or $20\%$ of the reference dimension, whichever is
larger. We define
\begin{equation}
    \mathrm{Match}
    \big(
        (\hat{d}_1,\hat{d}_2),(d_1,d_2)
    \big)
    =
    \bigwedge_{i=1}^{2}
    \left(
        |\hat{d}_i-d_i|
        \leq
        \delta(d_i)
    \right).
\end{equation}
The final size indicator is
\begin{equation}
\begin{split}
    \mathrm{size\_ok}
    =
    \mathbb{I}\Big[
    &\mathrm{Match}
    \big(
        (\hat{d}_1,\hat{d}_2),(d_1,d_2)
    \big)
    \\
    &\lor
    \mathrm{Match}
    \big(
        (\hat{d}_1,\hat{d}_2),(d_2,d_1)
    \big)
    \Big].
\end{split}
\end{equation}

\paragraph{Semantic Accuracy.}
The four indicators are equally weighted to obtain the report-level semantic
accuracy:
\begin{equation}
    \mathrm{SemAcc}
    =
    \frac{1}{4}
    \left(
        \mathrm{presence\_ok}
        +
        \mathrm{category\_ok}
        +
        \mathrm{location\_ok}
        +
        \mathrm{size\_ok}
    \right).
\end{equation}
The final SemAcc reported for a model is the average over all evaluated report
samples:
\begin{equation}
    \mathrm{SemAcc}_{\mathrm{final}}
    =
    \frac{1}{N}
    \sum_{n=1}^{N}
    \mathrm{SemAcc}^{(n)},
\end{equation}
where $N$ denotes the number of evaluated reports.

\subsection{Attribute-wise Analysis of Semantic Accuracy}
\label{sec:semacc_analysis}

\begin{table}[!ht]
\centering
\caption{
Detailed comparison of the four semantic attributes used in Semantic Accuracy
on UltraG-Bench. The best, second-best, and third-best results among all models
are highlighted in \rankone{green}, \ranktwo{blue}, and \rankthree{yellow},
respectively.
For UltraG-Agent, $I$, $Q$, and $L$ refer to InternVL3.5-8B, Qwen3-VL-8B,
and Lingshu-7B, while $US3$ and $S3$ refer to UltraSAM3 and SAM3, respectively.
}
\label{tab:semacc_breakdown}

\setlength{\tabcolsep}{6pt}
\renewcommand{\arraystretch}{1.1}
\scriptsize

\resizebox{0.8\linewidth}{!}{%
\begin{tabular}{lcccc}
\toprule
Model
& Presence$\uparrow$
& Category$\uparrow$
& Location$\uparrow$
& Size$\uparrow$ \\
\midrule

\multicolumn{5}{c}{\textit{Closed-source MLLMs}}\\
\midrule

Claude-4.5
& 0.8789
& 0.3871
& 0.0067
& 0.0001
\\

GPT-5.4
& 0.8139
& 0.4824
& 0.0305
& 0.0418
\\

\midrule
\multicolumn{5}{c}{\textit{Open-source General-purpose MLLMs}}\\
\midrule

Qwen3-VL-4B
& 0.5231
& 0.2328
& 0.0001
& 0.0065
\\

MiniCPM-V 2.6
& 0.7070
& 0.2978
& 0.0002
& 0.0020
\\

Qwen3-VL-8B
& 0.8173
& 0.2987
& 0.0005
& 0.0000
\\

Qwen3-VL-32B
& 0.8275
& 0.3223
& 0.0181
& 0.0007
\\

InternVL3.5-8B
& 0.8508
& 0.3678
& 0.0042
& 0.0025
\\

InternVL3.5-14B
& 0.9315
& 0.4092
& 0.0090
& 0.0113
\\

\midrule
\multicolumn{5}{c}{\textit{Medical MLLMs}}\\
\midrule

LLaVA-Med
& 0.8590
& 0.2416
& 0.0000
& 0.0002
\\

MedGemma-4B
& 0.8932
& 0.3484
& 0.0016
& 0.0003
\\

Lingshu-7B
& 0.7470
& 0.4121
& 0.0000
& 0.0000
\\

Lingshu-32B
& 0.6789
& 0.4836
& 0.0000
& 0.0054
\\

MedGemma-27B
& 0.8647
& 0.4418
& 0.0007
& 0.0037
\\

\midrule
\multicolumn{5}{c}{\textit{Pixel-level Grounded Medical MLLMs}}\\
\midrule

UniBiomed
& 0.2472
& 0.0646
& 0.0000
& 0.0003
\\

\midrule
\multicolumn{5}{c}{\textit{UltraG-Agent (Ours)}}\\
\midrule

UltraG-Agent$_{I+S3}$
& 0.1418
& 0.4068
& 0.0382
& 0.0000
\\

UltraG-Agent$_{L+US3}$
& \rankone{0.9614}
& \ranktwo{0.5853}
& \rankthree{0.5802}
& \ranktwo{0.4335}
\\

UltraG-Agent$_{Q+US3}$
& \rankthree{0.9500}
& \rankthree{0.5024}
& \ranktwo{0.5992}
& \rankthree{0.3997}
\\

UltraG-Agent$_{I+US3}$
& \ranktwo{0.9502}
& \rankone{0.7254}
& \rankone{0.6462}
& \rankone{0.4606}
\\

\bottomrule
\end{tabular}
}
\end{table}

Table~\ref{tab:semacc_breakdown} provides a detailed breakdown of Semantic Accuracy into its four constituent attributes: presence, category, location, and size. Existing MLLMs generally perform well on target presence, indicating that they can often recognize whether the relevant anatomical structure is visible in an ultrasound image. However, their performance decreases substantially for more fine-grained attributes. In particular, most general-purpose and medical MLLMs achieve very low accuracy on location and size, suggesting that semantic recognition does not necessarily translate into precise spatial understanding. Category recognition is comparatively stronger, but still remains substantially below the best UltraG-Agent variants.

UltraG-Agent shows clear improvements in the attributes that require pixel-level visual evidence. UltraG-Agent$_{L+US3}$ achieves the highest presence accuracy of 0.9614, while UltraG-Agent$_{I+US3}$ obtains the best results for category, location, and size, reaching 0.7254, 0.6462, and 0.4606, respectively. The gains are particularly pronounced for location and size, where conventional MLLMs achieve only limited performance. Moreover, replacing UltraSAM3 with the original SAM3 leads to substantial degradation across these attributes, further demonstrating the importance of ultrasound-specific pixel-level localization. Overall, the attribute-level results show that the improvement in SemAcc primarily comes from stronger alignment between semantic predictions and fine-grained spatial evidence, rather than from target recognition alone.

\subsection{Case Study}
\label{case_study}

\begin{figure}[!ht]
    \centering
    \includegraphics[width=\textwidth]{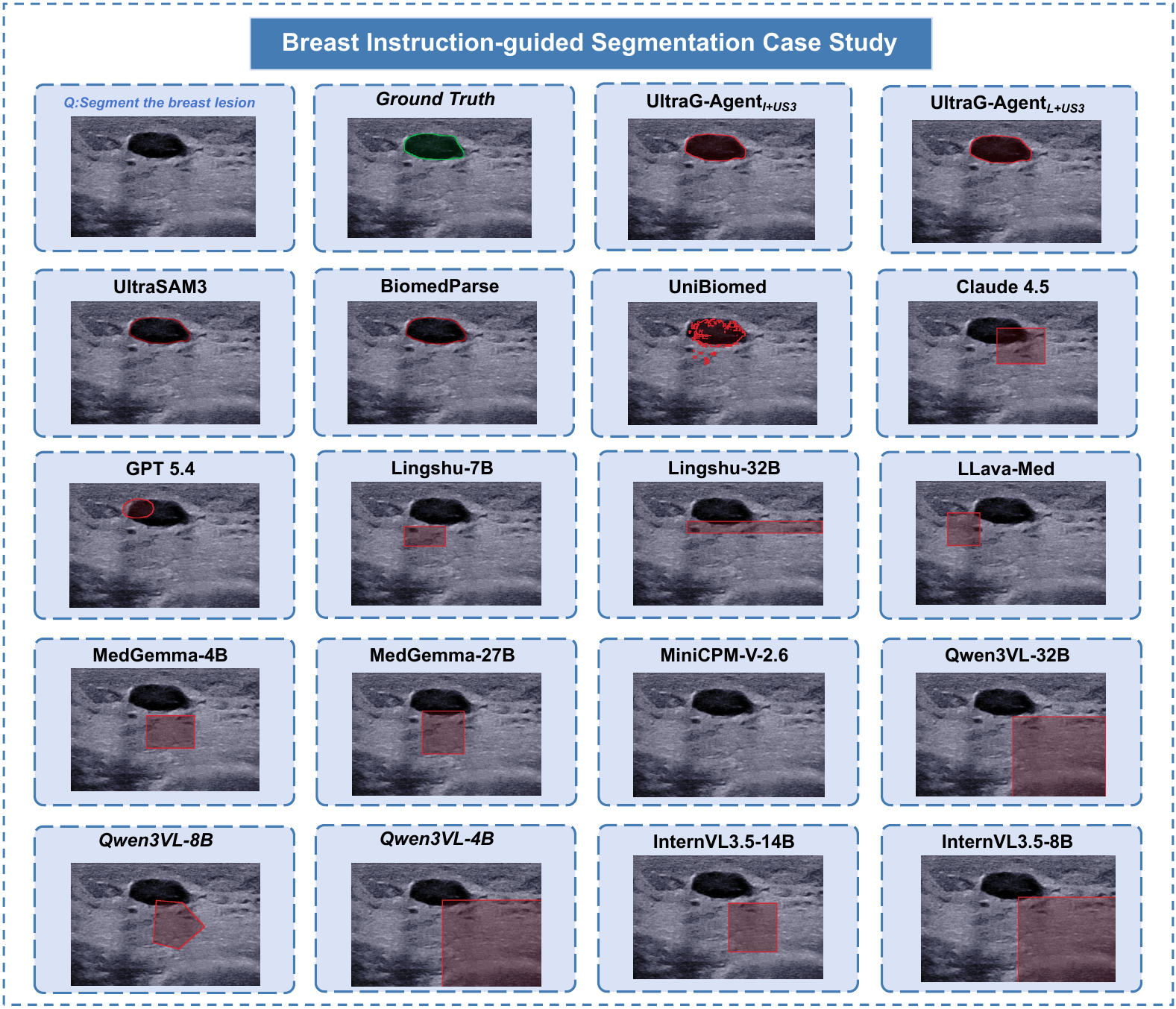}
    \caption{Breast Instruction-guided Segmentation Case Study.}
    \label{fig:cs_seg1}
\end{figure}

\paragraph{Instruction-guided Segmentation Case Study.}
Figure~\ref{fig:cs_seg1} presents a representative breast lesion segmentation example. UltraG-Agent produces pixel-level masks that closely follow the ground-truth lesion boundary, with UltraG-Agent$_{I+US3}$ showing particularly accurate spatial alignment and boundary delineation. UltraSAM3 and BiomedParse also achieve reasonable localization, reflecting the advantage of dedicated prompt-driven segmentation models for fine-grained spatial prediction. In contrast, most standalone MLLMs struggle to produce precise pixel-level outputs, often returning coarse bounding boxes, incomplete contours, or regions that only partially overlap with the target. This case illustrates the substantial gap between recognizing the requested anatomical target and accurately delineating its spatial extent.

\begin{figure}[!ht]
    \centering
    \includegraphics[width=\textwidth]{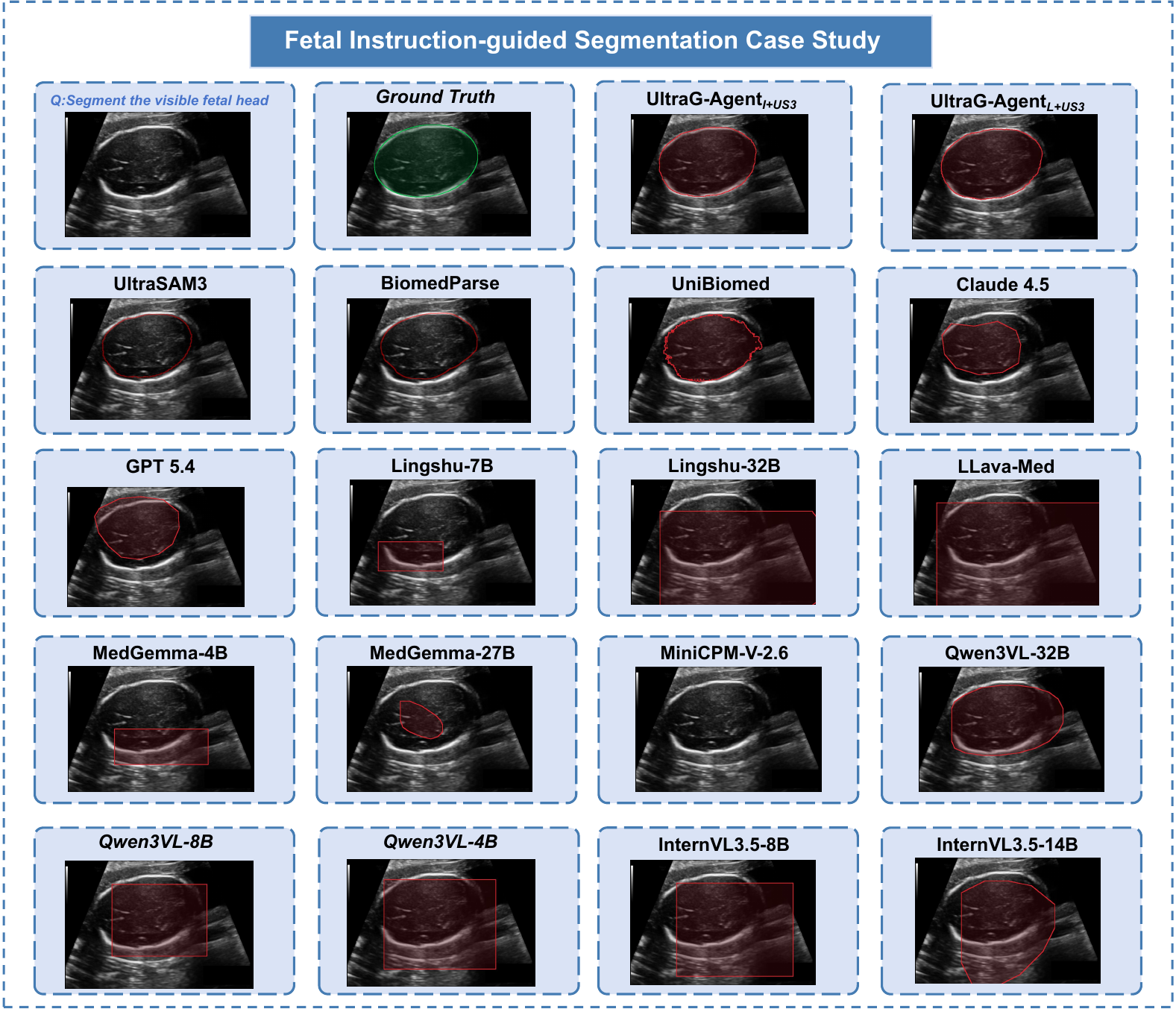}
    \caption{Fetal Instruction-guided Segmentation Case Study.}
    \label{fig:cs_seg2}
\end{figure}

Figure~\ref{fig:cs_seg2} provides another qualitative comparison under the same instruction-guided segmentation setting. A similar trend can be observed: UltraG-Agent generates a more complete and spatially consistent target mask, while many standalone MLLMs provide only approximate localization or fail to capture the target boundary accurately. Although prompt-driven segmentation models generally exhibit stronger localization ability than conventional MLLMs, UltraG-Agent retains this fine-grained segmentation capability while additionally supporting the semantic understanding required by the other tasks in UltraG-Bench. Together, the two cases demonstrate the complementary benefit of combining MLLM-based instruction understanding with ultrasound-specific pixel-level segmentation.

\begin{figure}[!ht]
    \centering
    \includegraphics[width=\textwidth]{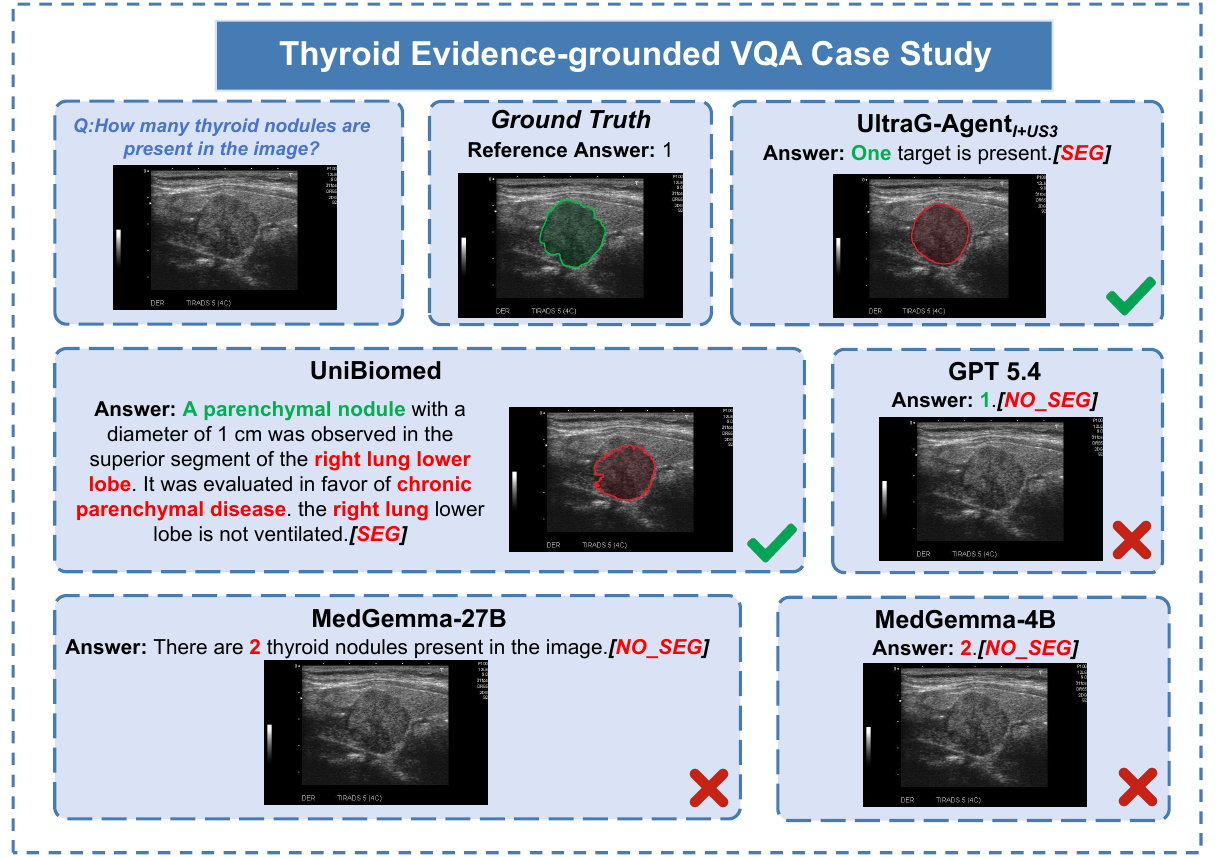}
    \caption{Thyroid Evidence-grounded VQA Case Study.}
    \label{fig:cs_vqa1}
\end{figure}

\paragraph{Evidence-grounded VQA Case Study.}
Figure~\ref{fig:cs_vqa1} presents a representative thyroid VQA example. UltraG-Agent correctly identifies that a single thyroid nodule is present and provides a well-aligned segmentation mask as pixel-level evidence. In contrast, several compared MLLMs either return an answer without valid grounding or generate incorrect and unsupported descriptions. This case demonstrates that a semantically plausible answer alone is insufficient without spatial evidence aligned with the target.

\begin{figure}[!ht]
    \centering
    \includegraphics[width=\textwidth]{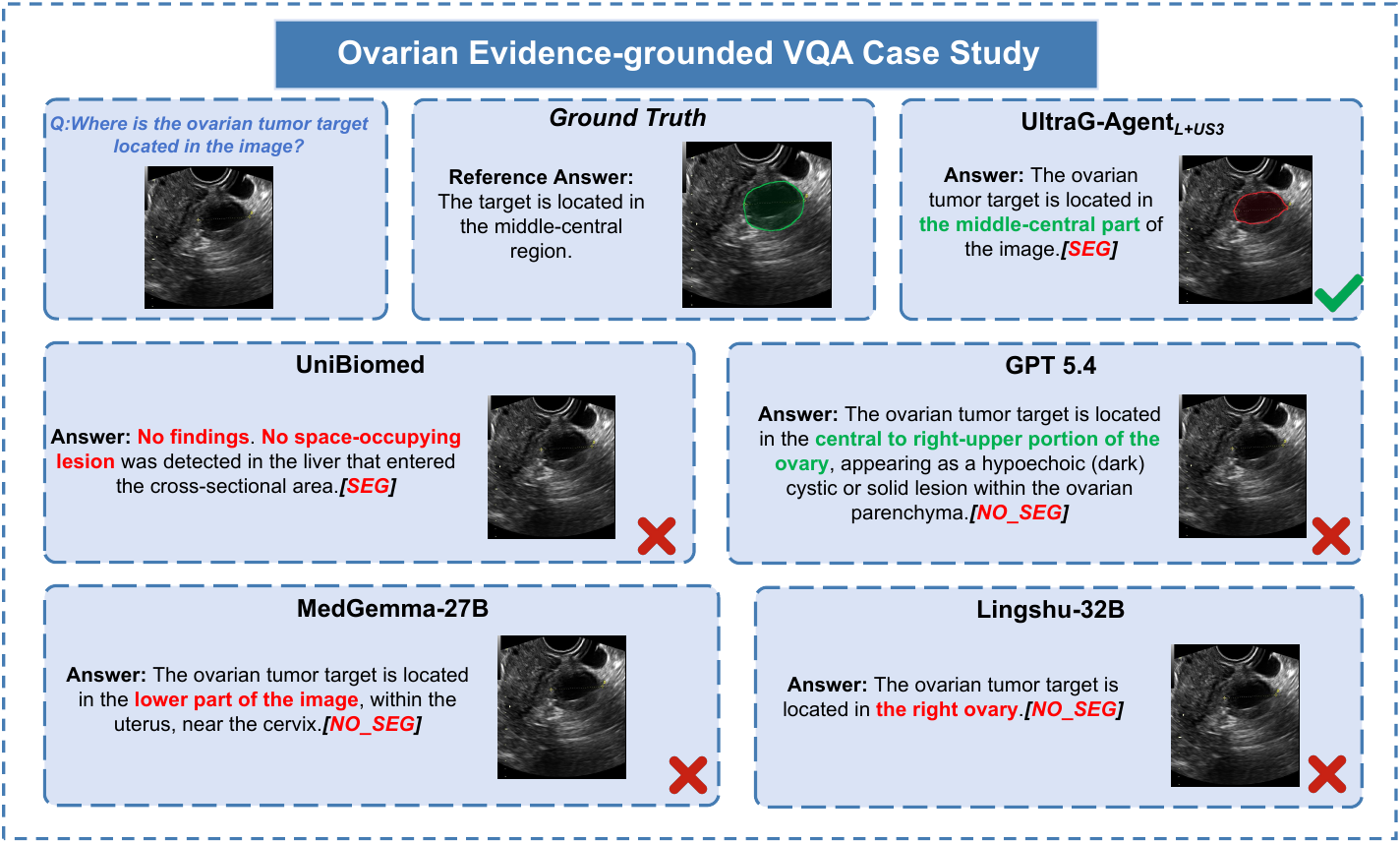}
    \caption{Ovarian Evidence-grounded VQA Case Study.}
    \label{fig:cs_vqa2}
\end{figure}

Figure~\ref{fig:cs_vqa2} shows a representative ovarian VQA example. UltraG-Agent correctly localizes the ovarian tumor target to the middle-central image region and grounds the answer with the corresponding segmentation mask. By comparison, the other models produce inaccurate spatial descriptions, unsupported anatomical references, or responses without valid pixel-level grounding. This case highlights the importance of explicitly aligning semantic answers with fine-grained visual evidence.

\begin{figure}[!ht]
    \centering
    \includegraphics[width=\textwidth]{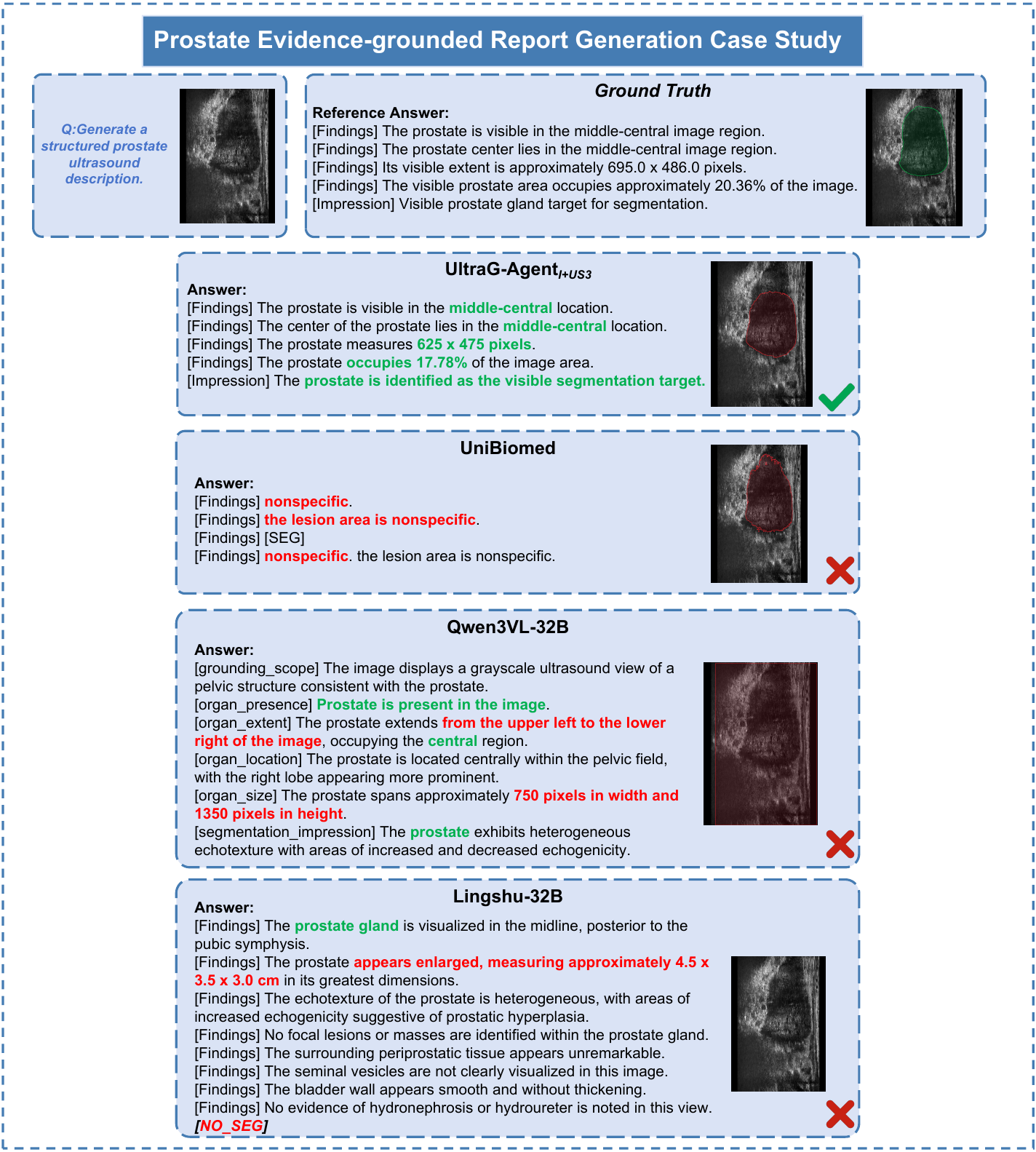}
    \caption{Prostate Evidence-grounded Report Generation Case Study.}
    \label{fig:cs_rg1}
\end{figure}

\paragraph{Evidence-grounded Report Generation Case Study.}
Figure~\ref{fig:cs_rg1} presents a representative prostate ultrasound report generation example. UltraG-Agent produces a structured report that closely follows the reference in target presence, middle-central location, pixel extent, and visible-area proportion, while providing a corresponding segmentation mask. In contrast, UniBiomed generates largely nonspecific statements despite producing a plausible mask, Qwen3-VL-32B introduces substantial errors in location and size estimation, and Lingshu-32B further hallucinates unsupported clinical attributes and physical measurements. This example shows that accurate report generation requires not only recognizing the target, but also grounding each spatial description in reliable pixel-level evidence.

\begin{figure}[!ht]
    \centering
    \includegraphics[width=\textwidth]{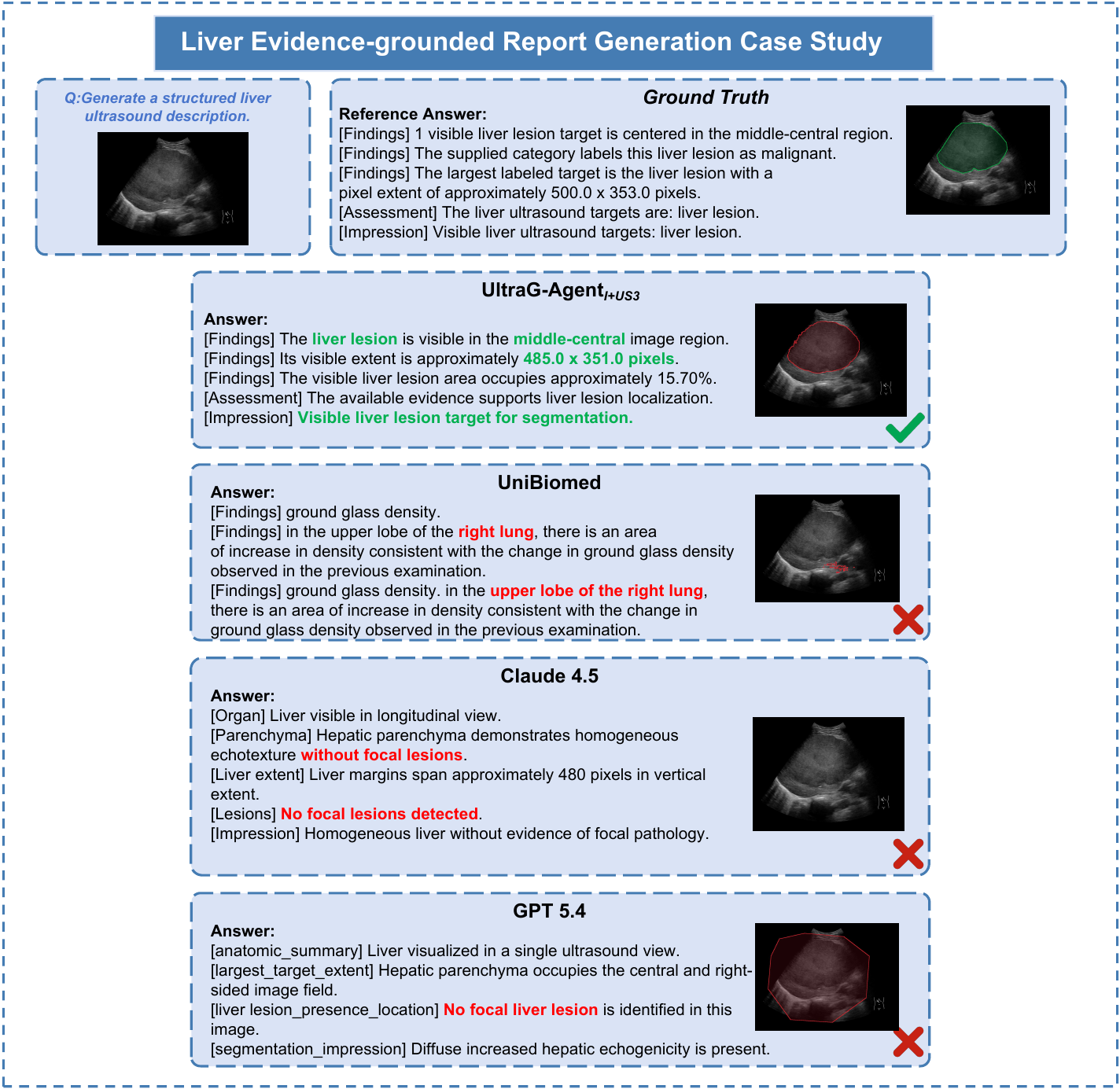}
    \caption{Liver Evidence-grounded Report Generation Case Study.}
    \label{fig:cs_rg2}
\end{figure}

Figure~\ref{fig:cs_rg2} shows a liver ultrasound report generation example. UltraG-Agent correctly identifies the liver lesion, localizes it to the middle-central image region, and reports its pixel-level extent and image-area proportion with a mask that closely matches the reference. By comparison, UniBiomed produces semantically incorrect content referring to lung findings, while Claude-4.5 and GPT-5.4 fail to identify the focal liver lesion and instead generate unsupported or contradictory descriptions. The case highlights the benefit of evidence-constrained generation, where report content is explicitly tied to the localized ultrasound target rather than generated from semantic cues alone.

\subsection{Error Analysis}
\label{error_analysis}

\begin{figure}[!ht]
    \centering
    \includegraphics[width=\textwidth]{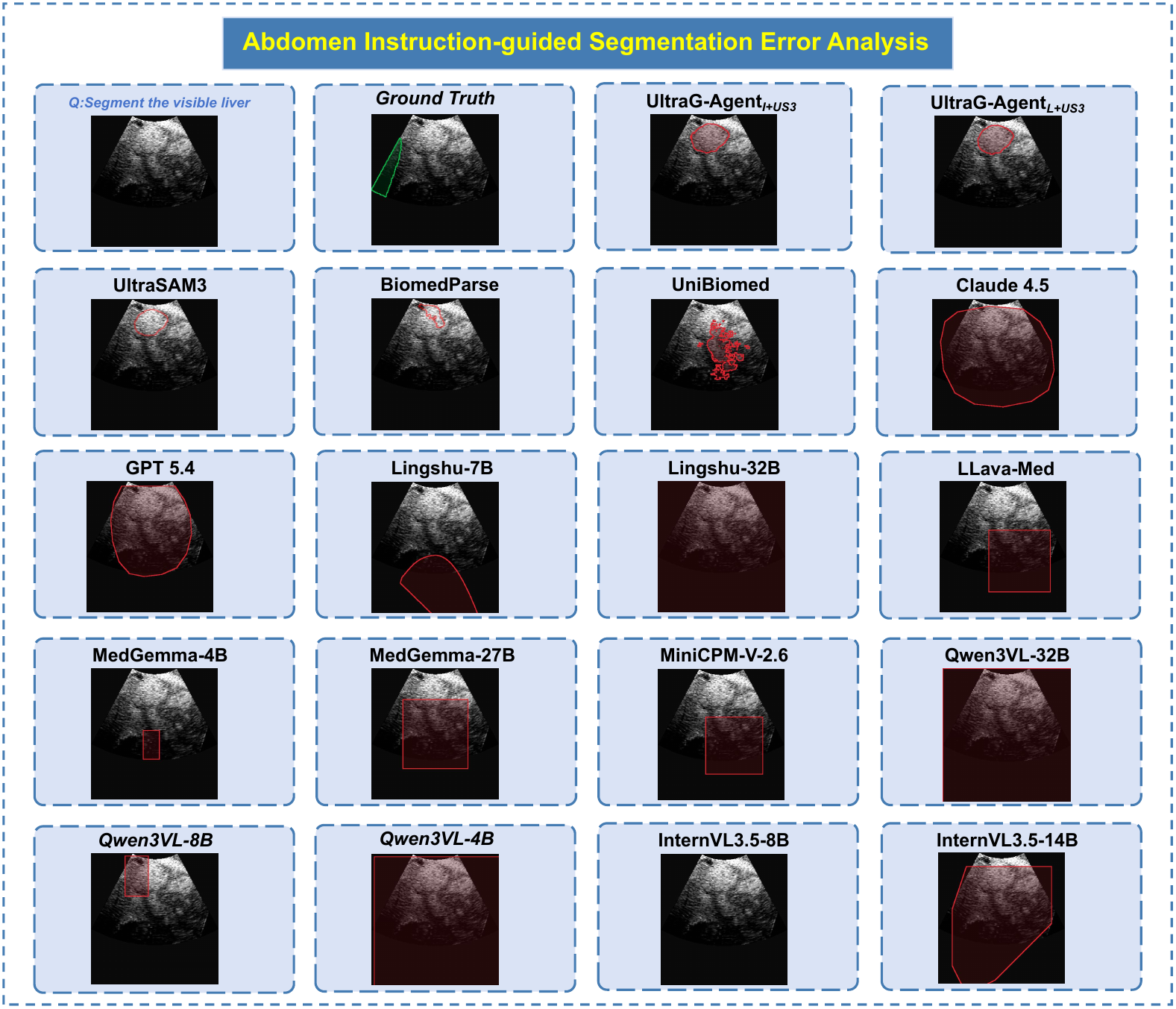}
    \caption{Abdomen Instruction-guided Segmentation Error Analysis.}
    \label{fig:ea_seg}
\end{figure}

\paragraph{Instruction-guided Segmentation Error Analysis.}
Figure~\ref{fig:ea_seg} illustrates a challenging abdominal ultrasound case in which most models fail to accurately delineate the visible liver region. UltraG-Agent and UltraSAM3 localize only a small region near the upper-central part of the image, resulting in a substantial mismatch with the ground-truth mask, while BiomedParse and UniBiomed produce incomplete or fragmented predictions. Standalone MLLMs exhibit more severe failure modes, including coarse bounding boxes, substantial over-segmentation, spatially shifted regions, and empty predictions. This case highlights the difficulty of segmenting targets with weakly defined boundaries and irregular visible extent in ultrasound. It also reveals a limitation of UltraG-Agent: when the underlying prompt-driven segmentation model misidentifies the target region, the localization error is propagated to the final grounded output, indicating that further improvements in ultrasound-specific segmentation remain important.

\begin{figure}[!ht]
    \centering
    \includegraphics[width=0.8\textwidth]{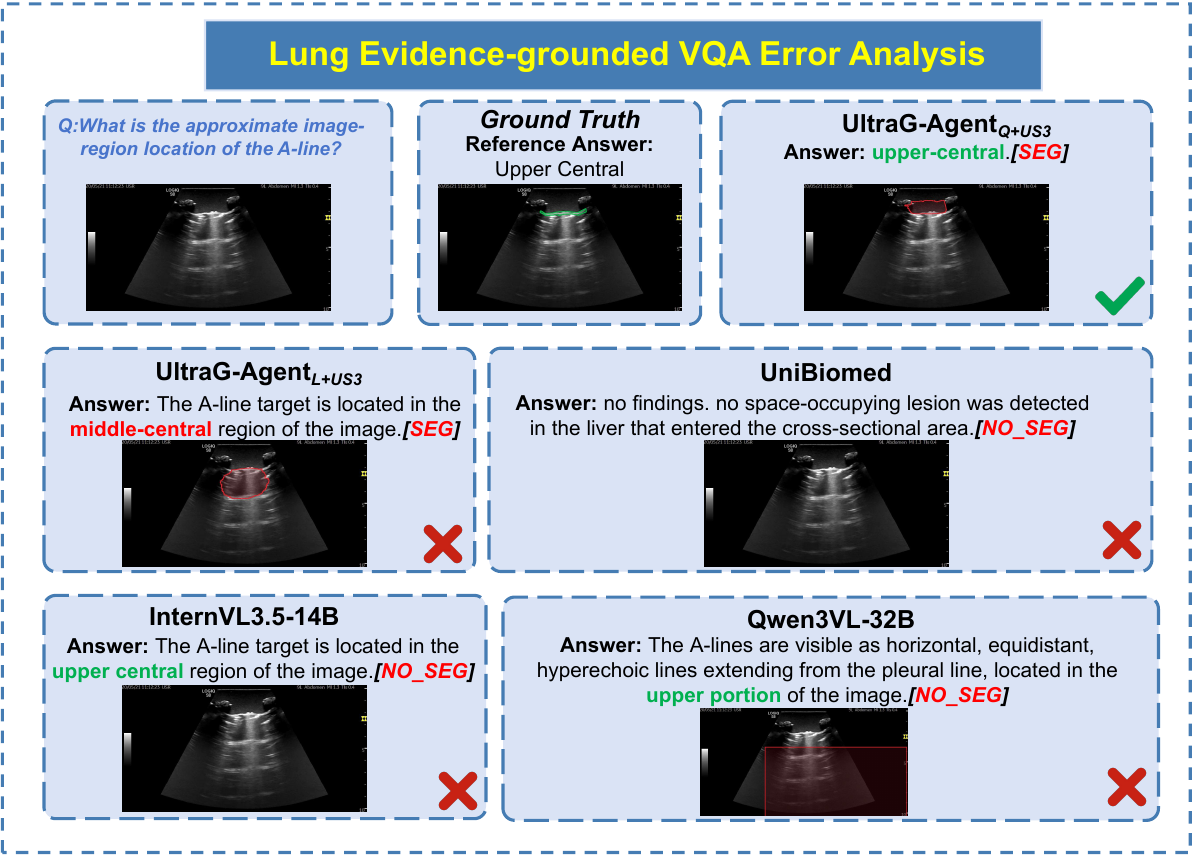}
    \caption{Lung Evidence-grounded VQA Error Analysis.}
    \label{fig:ea_vqa}
\end{figure}

\paragraph{Evidence-grounded VQA Error Analysis.}
Figure~\ref{fig:ea_vqa} presents a challenging lung ultrasound VQA case that exposes different failure modes in semantic prediction and visual grounding. UltraG-Agent$_{Q+US3}$ correctly identifies the A-line location as upper-central and provides a corresponding grounded region, while UltraG-Agent$_{L+US3}$ produces an incorrect middle-central localization despite generating a plausible mask. InternVL3.5-14B predicts the correct textual location but does not provide valid pixel-level evidence, illustrating that semantic correctness alone is insufficient for grounded VQA. UniBiomed fails to identify the relevant target, whereas Qwen3VL-32B generates an over-detailed description with inaccurate localization and an overly broad evidence region. This example highlights that evidence-grounded VQA requires both correct semantic reasoning and spatially consistent localization, and errors in either component lead to failure under the joint evaluation criterion.

\begin{figure}[!ht]
    \centering
    \includegraphics[width=\textwidth]{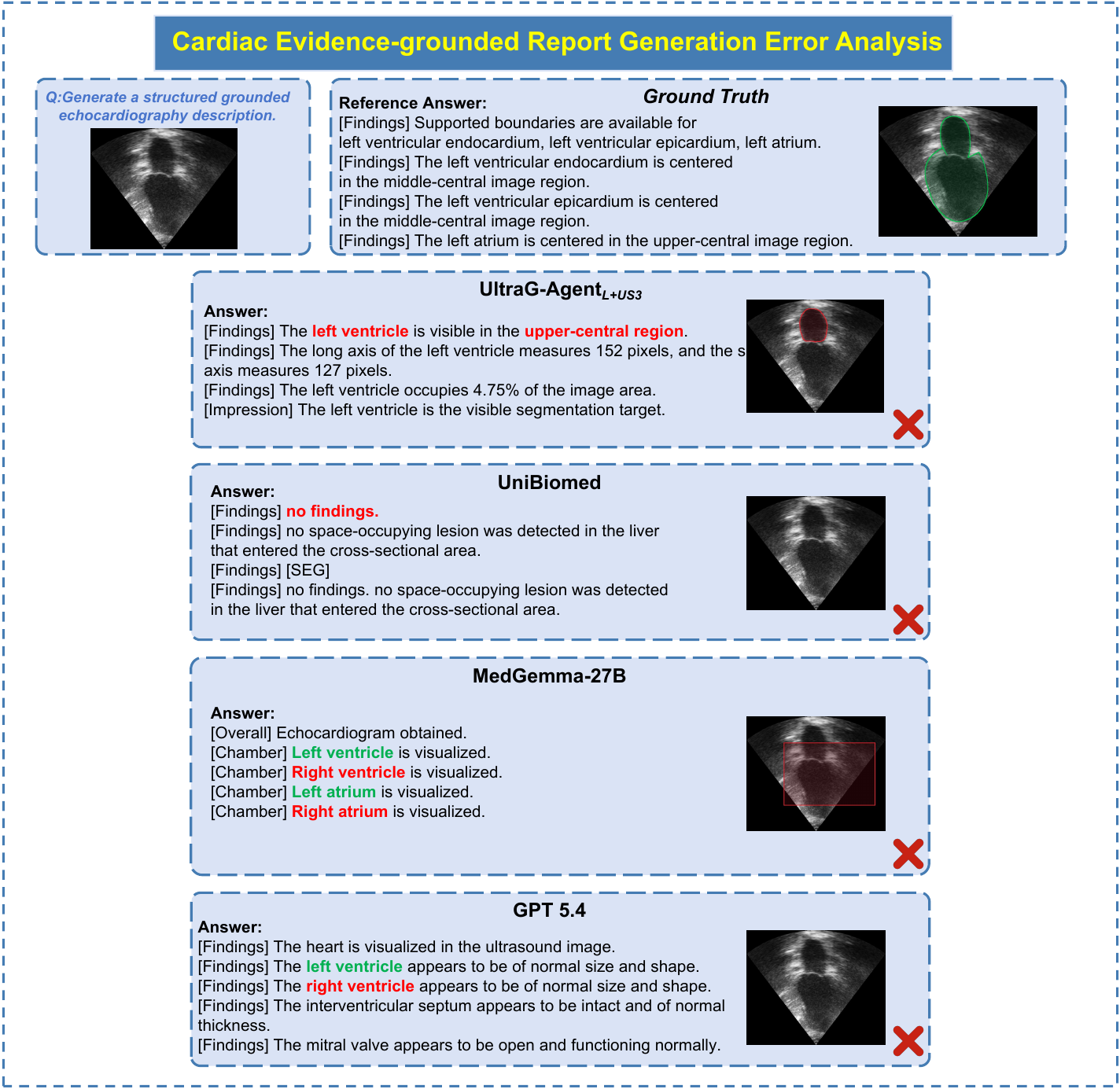}
    \caption{Cardiac Evidence-grounded Report Generation Error Analysis.}
    \label{fig:ea_rg}
\end{figure}

\paragraph{Evidence-grounded Report Generation Error Analysis.}
Figure~\ref{fig:ea_rg} presents a challenging cardiac ultrasound case involving multiple anatomical structures. The reference report requires grounded descriptions of the left ventricular endocardium, left ventricular epicardium, and left atrium, whereas UltraG-Agent identifies only a single left-ventricular target and further assigns an incorrect upper-central location. This illustrates a limitation of the current agent pipeline when an instruction requires simultaneous planning and grounding of multiple structures. UniBiomed fails to produce relevant cardiac findings, while MedGemma-27B recognizes several cardiac chambers but introduces unsupported structures and provides an imprecise evidence region. GPT-5.4 generates clinically plausible descriptions of cardiac anatomy, but adds unsupported attributes such as normal morphology and valve function without corresponding pixel-level evidence. Overall, this case shows that evidence-grounded report generation remains particularly challenging when multiple anatomical targets must be jointly identified, localized, and described, and that errors in target planning can propagate to the final grounded report.

\end{document}